\documentclass{article}

\usepackage{wrapfig}
\usepackage{microtype}
\usepackage{enumitem}
\usepackage{xcolor}
\usepackage{soul}
\usepackage{graphicx}
\usepackage{subfigure}
\usepackage{booktabs} % for professional tables
\usepackage{pifont}
\usepackage{xcolor}
\usepackage{booktabs}
\usepackage{multirow}
\usepackage{colortbl}
\usepackage{xcolor}
\usepackage{adjustbox}
\definecolor{darkgreen}{RGB}{0,120,0}
\definecolor{basebg}{RGB}{255, 245, 210}
\definecolor{oursbg}{RGB}{230,231,250}
\definecolor{gain}{RGB}{0,150,0}
\definecolor{loss}{RGB}{180,50,50}
\definecolor{darkred}{RGB}{139,0,0}
\definecolor{darkblue}{RGB}{0,0,139}
\usepackage{hyperref}

\usepackage[accepted]{icml2026}

\usepackage{amsmath}
\usepackage[most]{tcolorbox}
\usepackage{amssymb}
\usepackage{mathtools}
\usepackage{amsthm}
\usepackage{cuted}
\usepackage[capitalize,noabbrev]{cleveref}

\theoremstyle{plain}

\theoremstyle{definition}

\theoremstyle{remark}

\usepackage[textsize=tiny]{todonotes}

\icmltitlerunning{Mutual Information Preference Optimization}

\begin{document}

\twocolumn[
\icmltitle{Maximizing Mutual Information Between Prompt and Response Improves LLM Performance With No Additional Data}
% \icmltitle{Random data augmentation with DPO improves language model performance with no additional data}



\icmlsetsymbol{equal}{*}

\begin{icmlauthorlist}
\icmlauthor{Hyunji Nam}{yyy}
\icmlauthor{Haoran Li}{comp}
\icmlauthor{Natasha Jaques}{xxx}
\end{icmlauthorlist}

\icmlaffiliation{yyy}{Stanford University}
\icmlaffiliation{comp}{Character AI}
\icmlaffiliation{xxx}{University of Washington}

\icmlcorrespondingauthor{Hyunji Nam}{hjnam@stanford.edu}
\icmlcorrespondingauthor{Natasha Jaques}{nj@cs.washington.edu}

\icmlkeywords{Machine Learning, ICML}

\vskip 0.3in
]



\printAffiliationsAndNotice{} % otherwise use the standard text.

\begin{abstract}

While post-training has successfully improved large language models (LLMs) across a variety of domains, these gains heavily rely on human-labeled data or external verifiers. Existing data has already been exploited, and new data is expensive to collect. Moreover, true intelligence goes far beyond verifiable tasks. Therefore, we need self-improvement frameworks that are less dependent on external signals and more broadly applicable to both verifiable and non-verifiable domains. We propose \textbf{Mutual Information Preference Optimization (MIPO)}, a contrastive data augmentation method that constructs preference pairs by generating a positive response conditioning on the correct prompt, and a negative response by conditioning on a random, unrelated prompt. We show that using Direct Preference Optimization to learn from this paired data maximizes pointwise mutual information \emph{under the base LLM} between prompts and model responses. Experiments with with 1-7B parameter Llama and Qwen instruct models show that MIPO achieves 3--16\% gains (and 51\% increase for Qwen2.5-1.5B-Instruct) on personalization compared to prompting baselines. Surprisingly, MIPO can also be useful in verifiable domains, such as math and multiple-choice question answering, yielding 1--20\% gains \emph{without any additional data or external supervision}. These results suggest a promising direction for self-improvement using intrinsic signals derived from contrastive data pairs.

\end{abstract} 

% Surprisingly, we show that \emph{training with a model's own responses improves performance without additional data or reward functions}; even when the chosen data is potentially incorrect or suboptimal. 

\section{Introduction}
\emph{``Data is the fossil fuel of AI."} -- Ilya Sutskever

\begin{figure*}[h]
  \centering
  \includegraphics[width=\textwidth]{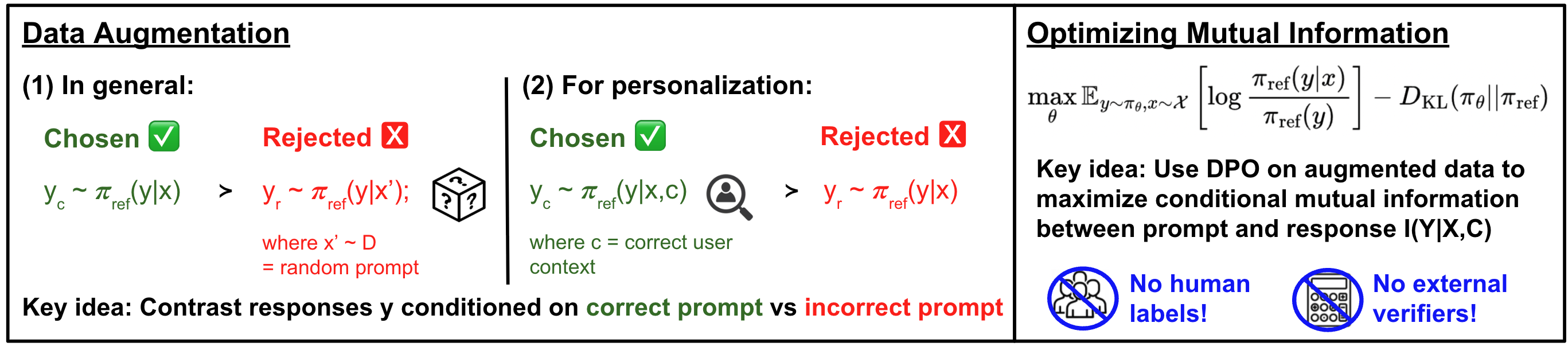}
   \vspace{-0.5cm}
      \caption{We propose an intrinsically motivated post-training method based on mutual information that does not require human labels or verifiers. We present two versions of MIPO: \textbf{(1) maximizing mutual information (under the base LLM) between responses and prompts}, and \textbf{(2) maximizing conditional mutual information between responses and user contexts given prompts} The second objective is particularly useful for personalization, as it encourages the policy to generate responses that are more likely conditioned on the specific user context, but rare globally.}
   \vspace{-0.5cm}
  \label{fig:intro_diagram}
\end{figure*}
Large language models (LLMs) have achieved remarkable success across diverse tasks and domains, from open-ended text generation to reasoning and mathematics. Post-training has been pivotal in driving this success through methods such as Reinforcement Learning with Human Feedback (RLHF)~\citep{stiennon2022learningsummarizehumanfeedback, touvron2023llamaopenefficientfoundation, ouyang2022traininglanguagemodelsfollow} and Reinforcement Learning with Verifiable Rewards (RLVR)~\citep{lambert2025tulu3pushingfrontiers, Guo_2025}. However, these methods still rely heavily on human feedback or external verifiers. As models become increasingly advanced and continue to develop beyond (average) human capabilities, this poses two challenges to existing frameworks: (1) intelligence extends far beyond verifiable tasks, and (2) high-quality human data is expensive to collect. The analogy of data as the ``fossil fuel" of AI applies not only to pre-training but also to post-training.

While alternative approaches, such as Reinforcement Learning from AI Feedback (RLAIF)~\citep{bai2022traininghelpfulharmlessassistant, lee2024rlaifvsrlhfscaling, chen2024iteraligniterativeconstitutionalalignment},  have been successful for reducing human supervision, self-training \emph{using the same model} without reliance on larger models remains under-explored. In fact, as a potential negative result,  ~\citet{huang2024largelanguagemodelsselfcorrect} observe that self-corrections by models can degrade performance when external feedback or verifiers are unavailable.

The lack of existing solutions for self improvement motivates our research: \emph{\textbf{Can models improve without external signals?}} Existing work shows some promise in this direction: even when the chosen responses are suboptimal, as long as they are better than the rejected counterparts (e.g., generated by a larger model), they can still lead to learning~\citep{yao2024varyingshadeswrongaligning, geng2025deltalearninghypothesispreference}. However, they still require a larger model to create supervised labels. If we think back to the literature on computer vision (for example, \citet{krizhevsky2012imagenet}), data augmentation approaches like translation and rotation improved learning without the use of any external supervision. Can we develop an analogous approach for LLMs, creating reliable learning signals intrinsic to the problem, without relying on human data or verifiers? 

% %We encounter two main challenges in pursuing this question. A key challenge is that the quality of a model’s own generated responses is hard to verify. Existing works have shown that we may not need to verify if a model's response is correct or not, if we can assume it is generally better than some alternative \citep{yao2024varyingshadeswrongaligning}. For example \citet{geng2025deltalearninghypothesispreference} show that creating preference pairs where an 8B model's response is always preferred to a 3B model's response can yield performance gains. 

We propose \textbf{Mutual Information Preference Optimization (MIPO)}, a data-augmentation method that maximizes mutual information between prompt $x$ and model response $y$ under the reference LLM. We show that the implicit reward in Direct Preference Optimization~\citep{rafailov2024directpreferenceoptimizationlanguage} can be cast as the InfoNCE~\citep{oord2019representationlearningcontrastivepredictive} objective, and we train the policy using chosen responses generated with the correct prompts $y_\text{chosen} \sim \pi_\text{ref}(y|x)$, and rejected responses generated with random, misspecified prompts $y_\text{rejected} \sim \pi_\text{ref}(y|x')$ for $x' \not= x$. 
% This induces the pointwise mutual information under the reference policy, $\log \frac{\pi_\text{ref}(y|x)}{\pi_\text{ref}(y)}$, acting as the implicit reward in DPO\footnote{With one negative per chosen response, the denominator is a single-sample Monte Carlo estimate of the marignal, $\pi_\text{ref}(y) = \mathbb E_x' [\pi_\text{ref}(y|x')]$, which we discuss in detail in the Methods section.}.

We apply this contrastive data technique to personalization problems, where models are given a prompt $x$ along with user-specific instructions or preferences $c$ to guide the personalized response. Context-driven personalization is increasingly becoming important given the growing need for pluralistic alignment~\citep{sorensen2024roadmappluralisticalignment}. Despite the growing capabilities of models, ~\citet{jiang2025artificialhivemindopenendedhomogeneity} alarmingly show the risks of model homogeneity, potentially ignoring unique user contexts and circumstances. We apply MIPO to maximize the relative likelihood of generating personalized responses $y_\text{chosen} \sim \pi_\text{ref}(y|x,c)$ compared to the rejected (un-personalized) responses $y_\text{rejected} \sim \pi_\text{ref}(y|x)$ or misspecified responses $y_\text{rejected} \sim \pi_\text{ref}(y|x, c')$. We show that MIPO encourages models to adapt to individual user contexts rather than relying on general information in the prompt. This also serves as a useful test bed for non-verifiable tasks that RLVR cannot effectively handle, which is a key aspect of our motivation.

Our empirical results with 1-7B parameter Llama and Qwen instruct models show that MIPO improves personalization by 3--16\% compared to personalized prompting baselines (Qwen2.5-1.5B-Instruct by 51\%). We evaluate five different models on one pluralistic benchmark (Multi-Bench~\citep{lee2024aligningthousandspreferencesmessage}) and two real-user datasets(PRISM~\citep{kirk2024prismalignmentdatasetparticipatory} and Community Alignment~\citep{zhang2025cultivatingpluralismalgorithmicmonoculture}). 

% \textcolor{blue}{METHOD --> personalization --> in-context steerability --> surprisingly, we find this approach can also work by randomizing prompts and achieve 90\% of the improvement by RLVR without any additonal labels / ground-truth answers}

% We then experiment with further extending MIPO to more general tasks beyond personalization, such as math and multiple-choice question answering, when there is no separation between queries and user contexts. This requires changing the training objective from the conditional mutual information between user contexts and model outputs given the queries, to the mutual information between prompts and model outputs directly. 
Surprisingly, MIPO also shows gains in verifiable domains \emph{without any verifiers or additional data}. We experiment on a suite of reasoning and MCQ tasks~\citep{gu2025olmesstandardlanguagemodel}, including GSM~\citep{cobbe2021trainingverifierssolvemath} and the AI2 Reasoning Challenge~\citep{clark2018thinksolvedquestionanswering}, which the models are likely to be already finetuned on. Yet we observe that MIPO, with no additional verifiable rewards, improves the performance of instruction-tuned models by 1--4\% on average across tasks, and by as much as 20\% for 1B Llama models. Furthermore, the performance gains achieved by MIPO often match or occasionally exceed RLVR which has access to privileged information. %and show that the benefits of MIPO can be generalized beyond personalization and are especially noticeable for smaller models (between 3 and 18\% improvements).  
% MMLU~\citep{hendrycks2021measuringmassivemultitasklanguage},

In summary, our contributions include:\begin{itemize}[noitemsep, topsep=0pt]
\item Proposing MIPO, a novel self-training method based on contrastive data augmentation and DPO;
\item Showing that the implicit reward in DPO is the pointwise mutual information between prompts and model outputs under the base LLM; 
\item Evaluating context-driven personalization on two real-user datasets and one personalized instruction-following benchmark, and showing that MIPO achieves 3--51\% empirical gains over the prompting baselines;
\item Applying MIPO to verifiable LLM benchmarks, such as GSM and the AI2 Reasoning Challenge, and showing an average improvement of 1--4\% on top of instruction fine-tuned models and up to 20\% for smaller models.
\end{itemize} 

\section{Related Work}
Compared to prior work on data construction for DPO~\citep{dwaracherla2024efficientexplorationllms, das2025activepreferenceoptimizationsample, gou2025mixedpreferenceoptimizationreinforcement, qi2025difficultybasedpreferencedataselection, yang2024weaktostrongreasoning, zhu2025weaktostrongpreferenceoptimizationstealing, yao2024varyingshadeswrongaligning, geng2025deltalearninghypothesispreference, xu2024automaticpairconstructioncontrastive, doosterlinck2024anchoredpreferenceoptimizationcontrastive}, we assume no external signals, stronger models, or ground-truth to guarantee the optimality of the chosen over the rejected responses. MIPO only requires that the chosen responses are generated from the correctly specified distribution, either by conditioning on the correct prompt or by conditioning on the correct prompt and context pair. Therefore, our work builds on self-training and self-improvement though many existing methods to this end also require verifiable rewards or human-supervised rubrics~\citep{dong2025stpselfplayllmtheorem, chen2024selfplayfinetuningconvertsweak, singh2024humandatascalingselftraining, ulmer2024bootstrappingllmbasedtaskorienteddialogue, liang2024isheepselfalignmentllmscratch, hosseini2024vstartrainingverifiersselftaught, yuan2023scalingrelationshiplearningmathematical}. ~\citet{yuan2025selfrewardinglanguagemodels, fränken2024selfsupervisedalignmentmutualinformation} are exceptions, as they use the same model to reward  the learned policy or generate training data. LLM personalization is also a growing field, but most existing methods rely on prompting or training with human (or a mix of real and synthetic) preferences~\citep{sorensen2024roadmappluralisticalignment, kirk2024prismalignmentdatasetparticipatory, zhang2025cultivatingpluralismalgorithmicmonoculture, sorensen2025spectrumtuningposttrainingdistributional,lee2024aligningthousandspreferencesmessage, kim2025cupidevaluatingpersonalizedcontextualized, zhao2025llmsrecognizepreferencesevaluating, jiang2025knowmerespondme, li2024personalizedlanguagemodelingpersonalized, liu2025sharedlowrankadaptationapproach, poddar2024personalizingreinforcementlearninghuman, nam2026learningsummarizeuserinformation}. In contrast, we do not require human supervision, and instead, focus on training models through intrinsic signals to effectively personalize responses based on the user-specific context given in a (context, prompt) pair. Due to space, we refer the reader to a more detailed discussion in Appendix~\ref{appendix:related_work}.

\section{Preliminaries}
\textbf{Post-training LLMs.} Let $\mathcal D$ be a dataset of annotated preferences consisting of a prompt $x$, and a corresponding pair of chosen and rejected responses to the same prompt $y_c$ (chosen) and $y_r$ (rejected). The Bradley Terry reward model $r$ assumes the following preference likelihood:
\begin{equation}
    p(y_c \succ y_r | x) = \frac{\exp r(x, y_c)}{\exp r(x, y_c) + \exp r(x,y_r)},
\end{equation} where the probability that $y_c$ is preferred over $y_r$ is a logistic function of the reward difference. The goal of preference optimization is to first learn this reward model $\hat r(x, y)$ using (human) labeled data, then learn a policy $\pi_\theta(y|x)$ that optimizes the learned reward subject to a KL-divergence constraint against the pre-trained reference policy $\pi_\text{ref}$. This leads to the following RLHF objective~\citep{jaques2017sequence, ouyang2022traininglanguagemodelsfollow}:  
\begin{equation}
  \max_{\pi_\theta} \mathbb E_{x \sim \mathcal D, y \sim \pi_\theta(.|x)} \left[\hat r(x,y) \right] - \beta \mathbb D_{KL}\left[\pi_\theta || \pi_\text{ref}\right].
\end{equation}

For personalized instruction-following~\footnote{We call it personalized instruction-following to clarify that the model is given user-specific context $c$ along with query $x$, and the goal is to generate a response about $x$ that follows the user's preference in $c$.}, we additionally define $c$ as the user-specific context, and modify the policy to condition on both the prompt $x$ and the context $c$, $\pi_\theta(y|x, c)$. Similarly, the reward is defined over both the prompt and the context as $r(x, y, c)$, consistent with prior work~\citep{poddar2024personalizingreinforcementlearninghuman, li2024personalizedlanguagemodelingpersonalized, nam2026learningsummarizeuserinformation}.

\textbf{Supervised Fine-tuning (SFT)} minimizes the forward KL divergence between the learned policy and the data generating policy~\citep{xiao2025connectionimitationlearningrlhf}, and the loss is computed over the chosen responses:

\begin{equation}
\mathcal L_\text{SFT}(\theta; \mathcal D) = - \mathbb E_{(x,y_c)\sim \mathcal D} \left[\log \pi_\theta(y_c|x)\right].
\end{equation}

\textbf{Direct Preference Optimization} (DPO)~\citep{rafailov2024directpreferenceoptimizationlanguage} is an alternative technique for learning from preference feedback that uses the same Bradley-Terry objective to directly modify the probabilities of the generator language model. 
The optimal policy $\pi_\theta$ is obtained from a reference model $\pi_\text{ref}$ via an energy re-weighting~\citep{lv2025hiddenlinkrlhfcontrastive}: 
\begin{equation}
\pi_\theta(y|x) = \pi_\text{ref}(y|x) \frac{\exp \left(\frac{1}{\beta} r(x, y) \right)}{Z(x)},
\end{equation} where $r$ is the reward, $\frac{1}{\beta} > 0$ is temperature, and $Z(x) = \mathbb E_{y \sim \pi_\text{ref}}\left[\exp \left(\frac{1}{\beta} r(x, y,)\right)\right]$. This simplifies the two stage RLHF process of finding the reward, then an optimal policy to maximize this reward, into a single step:
{\small
\begin{equation} \mathcal{L}_\text{DPO} = -\mathbb{E}_{\mathcal{D}} \left[ \log \sigma \left( \beta \log \frac{\pi_\theta(y_c|x)}{\pi_\text{ref}(y_c|x)} - \beta \log \frac{\pi_\theta(y_r|x)}{\pi_\text{ref}(y_r|x)} \right) \right] 
\end{equation}
}

% \begin{equation}
% \pi^*(y|x) = \frac{1}{Z(x)}\pi_{\text{ref}}(y|x) \exp\left(\frac{1}{\beta}r(x,y)\right),
% \end{equation} with $Z_\theta(x) = \sum_y \pi_\text{ref}(y|x)\exp \left(\frac{1}{\beta}r(x,y) \right)$ as the partition function, and incorporating (4) into the preference model in (1). This leads to the following loss:

\textbf{Contrastive Representation Learning} has been widely explored to learn latent representations that embed positive pairs close to each other and negative pairs far from each other. Most relevant to our work, \citet{oord2019representationlearningcontrastivepredictive} proposes the Information Noise-Contrastive Estimation (InfoNCE) loss, which ~\citet{poole2019variationalboundsmutualinformation} modifies as: \begin{equation}
\mathcal L_\text{infoNCE} = - \mathbb E_{\mathcal D} \left[\frac{1}{K}\sum_{i=1}^K \log \frac{\pi_\theta(y_i|x,c_i)}{\frac{1}{K}\sum_{j=1}^K \pi_\theta(y_i|x, c_j)} \right], 
\end{equation}
Minimizing this loss is equivalent to maximizing a lower bound on the mutual information between $X, Y$, where the positives are sampled from conditional distribution $p(y|x)$ and the negatives are sampled from marginal distribution $p(y)$. This objective has been adapted by numerous other works in RL and representation learning~\citep{chen2020simpleframeworkcontrastivelearning, eysenbach2023contrastivelearninggoalconditionedreinforcement, mazoure2022contrastivevaluelearningimplicit, hejna2024contrastivepreferencelearninglearning}. 

\citet{fränken2024selfsupervisedalignmentmutualinformation} extend InfoNCE to LLM settings in their method SAMI by maximizing the conditional mutual information between constitutions and responses, given by: \begin{multline}
\mathcal{L}_\text{SAMI} = - \mathbb{E}_{\mathcal{D}} \Biggl[ \frac{1}{2K} \sum_{i=1}^K \Biggl( \log \frac{\pi_\theta(y_i|x,c_i)}{\frac{1}{K}\sum_{j=1}^K \pi_\theta(y_i|x, c_j)} \\
+ \log \frac{\pi_\theta(y_i|x,c_i)}{\frac{1}{K}\sum_{l=1}^K \pi_\theta(y_l|x, c_i)} \Biggr) \Biggr]
\end{multline}

While SAMI and MIPO are both motivated by InfoNCE, SAMI only applies to settings where prompts and constitutions are decoupled and therefore cannot be used in verifiable benchmarks. In contrast, MIPO can flexibly incorporate different data augmentation techniques (generating from prompts alone, or from prompts and contexts) and is built directly on top of DPO, making it an easy plug-in pre-processing step before standard DPO training. 

\section{Mutual Information Preference Optimization (MIPO)} We propose two instantiations of MIPO based on the same underlying principle: maximizing the mutual information between prompts and model responses as an intrinsic reward signal. First, we introduce a simplified version that maximizes the mutual information between model outputs and prompts \emph{under the base policy} to make the connection with contrastive representation learning clear. Then we will modify this objective to maximize the conditional mutual information between responses and user contexts given the prompts.

\textbf{Setup:} For a given prompt $x \in \mathcal X$, we first sample a chosen response conditioned on the correct prompt: $y_c \sim \pi_\text{ref}(y|x)$, and pair it with a rejected response generated using a random prompt $y_r \sim \pi_\text{ref}(y|x'), x' \not= x$. We use this contrastive dataset of $(x, y_r, y_c)$ to train DPO.

% While this data augmentation process does not guarantee the absolute quality or correctness of positive responses $y_c$, this helps preserve their relative informativeness compared to the rejected responses, since the rejected samples are outputs to different prompts and therefore less likely to answer the original prompt.

\textbf{MIPO uses the pointwise mutual information under the base policy as an implicit reward.} DPO maximizes an implicit reward, replacing the two-step process used in traditional RLHF methods with single-step policy optimization. In typical RLHF settings, the training data comprises human-labeled preferences over chosen and rejected responses. However, in MIPO, \emph{what learning signal do randomly paired rejected responses provide relative to the chosen responses?} The key lies in DPO's connection to InfoNCE~\citep{oord2019representationlearningcontrastivepredictive}. InfoNCE maximizes a lower bound on the mutual information $I(X, Y)$:
\begin{equation}
I(X; Y) \geq \log(N) + \underbrace{\mathbb E_X \left[ \log \frac{ \exp r(x, y)}{\sum_{y_i \in \mathcal Y} \exp r(x, y_i)}\right]}_{- \mathcal L_{\text{infoNCE}}},
\end{equation}
where $\mathcal Y$ is a set of $N$ random samples containing one positive from the conditional distribution, $y_c \sim \pi_\text{ref}(y|x)$, and $N-1$ negative samples from the marginal, $y_r \sim \pi_\text{ref}(y)$.
Importantly, ~\citet{oord2019representationlearningcontrastivepredictive} (Section 2.3 \& A.1) show that the optimal critic is proportional to the density ratio of the conditional and the marginal distributions:
\begin{equation}
    r(x,y) \propto \log \frac{\pi_\text{ref}(y|x)}{\pi_\text{ref}(y)},
\end{equation} which is also the pointwise mutual information between $x, y$ under the reference model. In the case of one negative sample, InfoNCE becomes: \begin{equation}
\mathcal L_\text{infoNCE} = - \mathbb E_{\text (x, y_r, y_c) \sim \mathcal D}\left[\log \frac{\exp r(x,y_c)}{\exp r(x,y_r) + \exp r(x, y_c)} \right].
\end{equation} 
Rewriting Eq. (4) in terms of the reward yields: \begin{equation}
r(x, y) = \beta \log \frac{\pi_\theta(y|x)}{\pi_\text{ref}(y|x)} - \log Z(x),
\end{equation} and substituting this into the one-sample InfoNCE loss recovers the DPO objective. Therefore, when chosen responses are sampled according to $\pi_\text{ref}(y|x)$ and rejected responses are sampled from the marginal $\pi_\text{ref}(y)$\footnote{Since we cannot sample directly from the marginal, we use Monte Carlo approximation: sample random prompts $x'$, then generate $y_r \sim p(y|x')$. See Section 4.2 for a discussion of this approximation.}, DPO with MIPO's contrastive pairs optimizes the following objective:

{\small
\begin{equation}
\arg \max_{\pi_\theta} \mathbb E_{x \sim \mathcal X, y \sim \pi_\theta(y|x)} \underbrace{\left[ \log \frac{\pi_\text{ref}(y|x)}{\pi_\text{ref}(y)} \right]}_{\text{optimal reward } r(x,y)} - D_{\text{KL}}(\pi_\theta || \pi_\text{ref}).
\end{equation}} By optimizing this reward, we maximize the pointwise mutual information between prompts and model outputs under $\pi_\text{ref}$. This drives the learned policy $\pi_\theta$ to upweight responses likely under $\pi_\text{ref}(y|x)$ and downweight responses that are globally likely under the marginal $\pi_\text{ref}(y)$. Note that the reward is static and defined in terms of the base model $\pi_\text{ref}$, so the mutual information being maximized is with respect to the reference policy rather than the learned policy $\pi_\theta$. We hypothesize that this is actually a strength of MIPO. $\pi_\text{ref}$ provides a reliable model of natural language, so training the policy to maximize this fixed reward effectively steers $\pi_\theta$ to adapt to prompts. In contrast, using an on-policy version of mutual information may incentivize reward hacking by manipulating the learned policy's output distributions without affecting the underlying actions. To confirm this empirically, we conduct ablations with different approximations of mutual information under $\pi_\theta$ (PPO) and $\pi_\text{ref}$ (DPO).

\subsection{Maximizing Conditional Mutual Information of Contexts and Responses Given Prompts} We use the mutual information between responses $y$ and user contexts $c$ conditioned on the prompts $x$ to learn an optimal user-conditioned policy $\pi_\theta(y|x,c)$. For example, given a user's query $x$ = \emph{``Explain random variables''}, there could be two different user contexts. $c$ = \emph{``I have a math PhD''}, and  $c'$ = \emph{``I am in 7th grade and just learned about probability''}. The goal of personalization is not only to respond to the prompt $x$, but also to address the user-specific information captured in $c$. This changes the mutual information in Eq. (8) to the following conditional objective:

{\small
\begin{equation}
I(Y; C | X) \geq \mathbb E_X \left[ \log \frac{ \exp r(x, y_c, c)}{\sum_{y_i \in \mathcal Y} \exp r(x, y_c, c) + \exp r(x, y_r, c)}\right].
\end{equation}
} This yields the following optimal critic: \begin{equation}
r^*(x, y, c) \propto \log \frac{\pi_\text{ref}(y|x,c)}{\pi_\text{ref}(y|x)}.
\end{equation} Maximizing the conditional pointwise mutual information $I_\text{ref}(y;c|x)$ improves the model's steerability by encouraging $\pi_\theta(y|x,c)$ to focus on user-specific context $c$ rather than relying on generic information in $x$ that works for any user. 
% Note that unlike the previous case, the negatives can be sampled directly from $\pi_\text{ref}(y|x)$ without further approximation, i.e., by sampling from the reference model conditioned on the prompt alone.

\subsection{Different Negative Sampling Approaches} While chosen responses are always generated with the correct prompt $x$ (or correct prompt and context $(c, x)$), we consider different strategies for sampling the negatives to pair with each positive response. We conduct ablations with (i) missing, (ii) random and (iii) reshuffling for personalized instruction-following, and (i) random and (ii) reshuffling for verifiable tasks that do not decouple prompts and contexts.

\textbf{When prompts and contexts are distinguished, negatives can be generated in the following two ways}:
\[
y_r \sim
\begin{cases}
\pi_\text{ref}(y|x) & \text{(i) \emph{missing} context} \\
\pi_\text{ref}(y|x, c'), c' \sim \mathcal C & \text{(ii) \textit{random} context}  \\
\end{cases}
\]
Strategy (i) samples directly from the prompt only, without the context. Strategy (ii) first samples a random user context $c'$, then generates a response conditioned on the prompt and the mismatched context $c'$. Since $\pi_\text{ref}(y|x) = \mathbb E_{c'} [\pi_\text{ref}(y|x,c')]$, Strategy (ii) is a unbiased estimator of $\pi_\text{ref}(y|x)$, but each negative sample $y_r$ is conditioned on $c'$. The second approach presents a trade-off. On the one hand, this may sharpen the contrast between chosen and rejected responses, since the rejected response is anchored to a different context. However, conditioning on the additional context introduces approximation error that vanishes only in expectation over many sampled contexts, so using a single negative may provide a coarse approximation.

\textbf{When prompts and contexts are not decoupled, negatives cannot be directly sampled from the marginal.} Since we do not have access to the marginal distribution $\pi_\text{ref}(y)$, we approximate sampling from the true marginal with $\mathbb E_{x' \sim \mathcal X} [\pi_\text{ref}(y|x)]$. We first sample a random prompt $x'$, then sample $y_r \sim \pi_\text{ref}(y|x')$. Pairing each positive sample with multiple negatives is more desirable than a single negative, since averaging across many $x'$s reduces the approximation error from conditioning on a particular $x'$. We conduct ablations with different numbers of negatives to test whether theory holds in practice, though we expect learning gains to plateau after some point.

We propose reshuffling the chosen responses as a practical alternative to re-generating with random prompts. While reshuffling violates the MIPO assumption that chosen and rejected responses are sampled independently from their respective distributions, it substantially reduces data-generation costs, which makes it appealing in practice. This also provides a clean comparison to SFT baselines, since SFT and MIPO (with reshuffling) use the exact same training data. Both use the same chosen responses, but MIPO takes the additional step of reshuffling them to serve as rejected responses paired with the originals. This allows us to empirically isolate the contribution of the contrastive signal from the benefit of additional training data, since if the rejected responses were generated independently, MIPO would have $2\times$ the training data.

\section{Experiments}

\subsection{Baselines}
\textbf{Personalized Prompting.} For personalized instruction-following, models are prompted with additional user context $c$ along with prompt $x$ to personalize their responses. The default baselines are only conditioned on prompts $x$. 

\textbf{Supervised Fine-Tuning (SFT)} is implemented in two ways: SFT without revision and SFT with revision. SFT without revision uses the same training data as MIPO (reshuffling). SFT with revision uses additional information to refine the model's initial answers. \textbf{Revision based on privileged information} (in verifiable tasks) uses ground-truth solutions: models are given the correct final answers and asked to revise their original answers accordingly. This privileged information is not available to MIPO or other unsupervised methods. \textbf{Revision based on contrastive answers} (in open-ended domains) improves the model's initial (chosen) response by comparing it to the rejected response and refining it (see the prompts for self-revision in Appendix \ref{appendix:sft_revision}). The rejected response used for comparison is generated with the same prompt but a random context.
% The prompt contains a list of user-specific attributes (also available during the initial generation), a \emph{good} response generated from the correct context, and a \emph{bad} response generated from a mismatched context with the same prompt. 
% We consider two versions of revision based on different degrees of privileged information made available to the model: (1) revision based on comparisons and (2) revision based on ground truth. In (1), models are given two responses (one generated from the correct prompt and context, and the other generated from the correct prompt but an incorrect context) and are prompted to improve the first response while avoiding mistakes observed in the second. In (2), which is only applicable to verifiable settings, models are provided with the correct answer along with their initial attempt and asked to revise their response. 

% \textbf{PPO~\citep{schulman2017proximalpolicyoptimizationalgorithms}} is used for RLVR and RLAIF training. 

\textbf{RL with Verifiable Rewards (RLVR)} is only applicable in verifiable tasks. We use exact string matching on the extracted model response to reward the correct response by 1 and 0 otherwise. RLVR uses privileged information not given to MIPO, so we expect it to outperform other methods including MIPO. We are interested in how much of the gains can be matched by MIPO without any verifier signal. 

\textbf{RL with AI Feedback} (RLAIF)~\citep{bai2022traininghelpfulharmlessassistant, lee2024rlaifvsrlhfscaling} uses LLM-as-a-judge to assign rewards. The judge receives the user's query and context along with the model's response to give a score between 1--5 based on how well the response addresses the user-specific context. For verifiable tasks, we provide the judge with the ground-truth answer to score the model's reasoning and answer on a scale of 1--5. Rather than using a stronger critic model, we use the same, but frozen, model as the critic to enable a fair comparison, since no other method is trained with signals from a larger model. RLVR and RLAIF are both optimized with PPO.

\subsection{Mutual Information Alternatives} 
For personalized instruction-following, where contexts and prompts are decoupled, we compare MIPO to other methods based on conditional mutual information. Specifically, we implement the following baselines: 

\textbf{(i) MI-PPO} uses PPO to maximize the pointwise conditional mutual information under the learned policy as:
\begin{equation}
r_{\text{MI}}(x, y, c) = \log \pi_\theta(y|x,c) - \log \pi_\theta(y|x).
\end{equation} \textbf{(ii) MI-RLAIF} combines mutual information reward with the task-specific reward from LLM-as-a-judge as follows: \begin{equation}
r_\text{MI\_RLAIF}(x, y, c) = r_\text{LLM\_as\_judge}(x, y, c) + \alpha r_\text{MI}.\end{equation}\textbf{(iii) InfoNCE}~\citep{oord2019representationlearningcontrastivepredictive}, and \textbf{(iv) SAMI}~\citep{fränken2024selfsupervisedalignmentmutualinformation, poole2019variationalboundsmutualinformation} are different instantiations of contrastive learning based on mutual information. We extend the original InfoNCE loss to LLM settings. SAMI (Eq. (7)) is closely related to InfoNCE but uses two symmetric loss terms: one contrasting different contexts for a fixed response, and the other contrasting different responses for a fixed context.

% \begin{itemize}
% \item SFT (using only the chosen responses of DPO dataset, i.e., model's initial self-generation)
% \item SFT with supervised self-revision (Revised SFT)
%     \begin{itemize}
%     \item two forms of supervision: (1) weak supervision for non rlvr domains: have response A and response B and the prompt $\rightarrow$ prompt the model to further improve A while not making the mistake of B
%     \item (2) strong supervision for rlvr: have response A + correct answer $\rightarrow$  prompt the model to improve its reasoning / solution to get the correct answer
%     \end{itemize}
% \item DPO with different amounts of misspecification in rejected samples (include the following as ablations on a few of the domains)
%     \begin{itemize}
%         \item random context but same prompt
%         \item random prompt
%         \item different domain (e.g., free response question but the response is generated for a math problem)
%         \item gibberish alphanumeric strings of the same length
%     \end{itemize}
% \item DPO with oracle chosen and rejected (as the upper bar; available for prism, community alignment, rlvr settings)
% \item think of other DPO baselines from prior works
% \item PPO with AI feedback (LLM-as-a-judge)
% \end{itemize}

% (i.e., potential gap between self-improvement and training with external feedback). 

\textbf{Implementation.} Our experiments are implemented using OpenRLHF~\citep{hu2024openrlhfeasytousescalablehighperformance} and conducted with five different models (Qwen2.5-1.5/3/7B-Instruct~\citep{qwen2025qwen25technicalreport} and Llama-3.2-1/3B-Instruct~\citep{llama3.2}). Training details are in Appendix~\ref{appendix:training} and our code is available at \url{https://github.com/nam630/mutual_information_preference_optimization}.
% Bigger models are used to evaluate text generation, e.g., Llama-3-8B-Instruct~\citep{grattafiori2024llama3herdmodels})

\subsection{Domains} We evaluate on 7 tasks including personalized instruction-following {\includegraphics[height=1em]{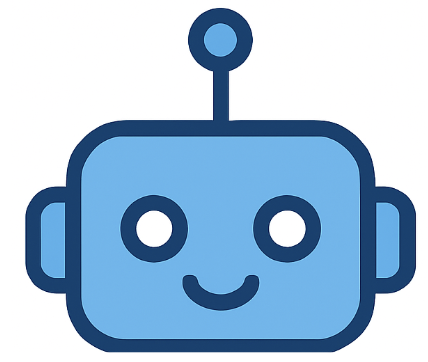}}, math benchmarks {\includegraphics[height=1em]{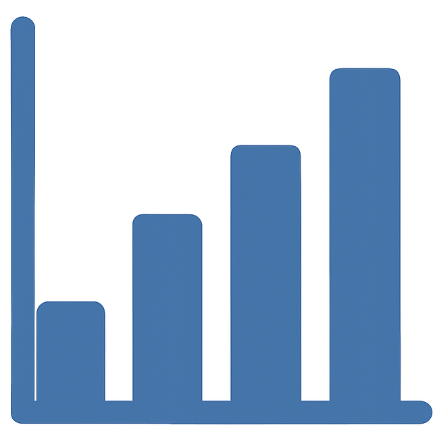}} (GSM8k~\citep{cobbe2021trainingverifierssolvemath} and  SVAMP~\citep{patel2021nlpmodelsreallyable}), and multiple-choice question answering (MCQ) (AI2 Reasoning Challenge (ARC)~\citep{clark2018thinksolvedquestionanswering}). For real-user datasets, we need to preprocess the data to obtain (context, query) pairs for training and evaluation. This helps simulate a setting where the user asks a general query along with specific requests or contexts, and the model needs to personalize the response. Details on the personalization tasks follow, and the LLM benchmarks are described further in Appendix~\ref{appendix:datasets}. We report final-answer accuracy on math and MCQs.

\textbf{{\includegraphics[height=1em]{llm_clipart.png}} Community alignment~\citep{zhang2025cultivatingpluralismalgorithmicmonoculture}} is a large-scale dataset collected from 3,000 participants across five countries, designed to capture pluralistic preferences and values. Since users select chosen and rejected responses without explicitly stating their ground-truth preferences, we use GPT-4o to infer the underlying preferences as user contexts for both training and testing. We sample 6,564 unique prompt-user pairs for training and 200 for evaluation.
% We then evaluate whether reward models conditioned on this user information can accurately recover ground-truth preferences (Qwen2.5-7B-Instruct achieves 98.5\%, and Meta-Llama-3-8B-Instruct achieves 92.0\%). Model responses are compared to the chosen responses from the dataset. The training set size is 6,564 and the test set size is 200.

\textbf{{\includegraphics[height=1em]{llm_clipart.png}} PRISM~\cite{kirk2024prismalignmentdatasetparticipatory}} is a pluralistic dataset comprising open-ended conversations from 1,500 users across 75 countries and 20 LLMs. As above, we use GPT-4o to infer user contexts from ground-truth chosen and rejected responses. The models are tested on how well they address the user-specific context when given a (context, query) pair. We sample 7,294 data points for training and 194 for testing.
% and validate the reward model's accuracy conditioned on these attributes (Qwen2.5-7B-Instruct achieves 92.6\%, and Meta-Llama-3-8B-Instruct achieves 92.0\%). Model responses are evaluated against the ground-truth chosen responses based on how well they adhere to user-specific instructions and values. The training set size is 7,294 and the test set size is 200.

\textbf{{\includegraphics[height=1em]{llm_clipart.png}} Multifaceted Bench~\citep{lee2024aligningthousandspreferencesmessage}} is a benchmark for evaluating models' ability to generate context-specific responses. It includes 921 instructions from five existing benchmarks (AlpacaEval 2.0~\citep{dubois2025lengthcontrolledalpacaevalsimpleway}, FLASK~\citep{flask}, Koala~\citep{koala}, MT-Bench~\citep{mtbench}, and Self-Instruct~\citep{selfinstruct}), each paired with synthetic, human-verified system messages specifying user preferences along four dimensions: style, background knowledge, informativeness, and harmlessness. We use GPT-4o to rewrite each system message into four first-person user messages, one per preference dimension, and split the dataset into 3,600 prompt-context pairs for training and 84 for testing. 

\subsection{Personalization Evaluation with LLM-judges {\includegraphics[height=1em]{llm_clipart.png}}} We use larger LLMs from the same model family (Qwen2.5-14B-Instruct for the Qwen models, and Llama-3-8B-Instruct for the Llama models) as judges to score responses (see LLM-judge prompts in Appendix~\ref{appendix:reward_prompt}). To validate the judges' alignment with human preferences, we measure the tie-tolerant accuracy of predicting ground-truth human preferences from PRISM and Community Alignment: 8B Llama achieves 92\% accuracy on both, and 14B Qwen achieves 93.5\% and 92.2\%, respectively. In our main results section, we report tie-inclusive win rates of the model-generated responses against ground-truth chosen responses (PRISM and Community Alignment datasets) and against personalized prompted GPT-4o responses (Multifaceted Bench). 

\textbf{Validation with diverse LLM judges.} In additional experiments using a larger 70B model, we evaluate responses across a suite of stronger proprietary LLMs (GPT-4.1, GPT-5.2, Gemini 2.5 Flash, and Gemini 3.1 Flash Lite Preview). In the 70B experiment setup, each judge is shown two candidate responses (one from MIPO and the other from personalized prompting) and asked to select the preferred response according to the personalization rubric (see the LLM-judge prompt in Appendix~\ref{appendix:binary_evaluation}).

% \textbf{General problem solving benchmarks} {\includegraphics[height=1em]{math_clipart.png}} {\includegraphics[height=1em]{alphabet_clipart.png}}.  We report the accuracy of the final answer on math {\includegraphics[height=1em]{math_clipart.png}} and MCQ {\includegraphics[height=1em]{alphabet_clipart.png}}. 
% % \vspace{-0.2cm}

\section{Results}
\textbf{R1: MIPO improves personalized instruction-following by 3--51\%, on average across datasets, over prompting.} 

\begin{table}[h!]
% \centering
% \begin{minipage}{0.5\textwidth}
% \vspace{-4cm}
\caption{{\includegraphics[height=1em]{llm_clipart.png}}  \textbf{Personalization win-rates evaluated on (i) Community Alignment (CA)~\citep{zhang2025cultivatingpluralismalgorithmicmonoculture}, (ii) PRISM~\citep{kirk2024prismalignmentdatasetparticipatory}, (iii) Multi-Bench (MB)~\citep{lee2024aligningthousandspreferencesmessage}.} Reported mean $\pm$ std from 3 seeds. Changes larger than 3 points from personalized prompting are in \textcolor{blue}{blue (for increase)} and \textcolor{darkred}{red (for decrease)}. We report the best number from different negative sampling strategies with MIPO.} 
\label{tab:personalization}
\centering
\adjustbox{width=0.5\textwidth}{
\begin{tabular}{lccc}
\toprule
\textbf{Model} & \textbf{CA} & \textbf{PRISM} & \textbf{MB} \\
\midrule
\textbf{Llama-3.2-1B-Instruct} & 31 & 36.6 & 59.52 \\
\rowcolor{basebg}
\quad + Personalized Prompting & 78.00 & 72.17 & 79.76  \\ 
\quad + SFT & \textcolor{blue}{83.83 $\pm$ 1.53} & 69.76 $\pm$ 1.07 & 78.97 $\pm$ 3.0 \\
\quad + SFT (Revision) & 80.00 $\pm$ 1.00 & 73.37 $\pm$ 2.15 & 77.78 $\pm$ 3.00 \\
\midrule
\quad + PPO (MI) & 80.17 $\pm$ 3.33 & 73.37 $\pm$ 5.86 & 82.22 $\pm$ 4.05 \\
\quad + RLAIF & 79.83 $\pm$ 5.69 &  \textcolor{darkred}{56.87 $\pm$ 2.44} & \textcolor{blue}{84.12 $\pm$ 0.69} \\
\quad + RLAIF (MI) & \textcolor{blue}{81 $\pm$ 2.50} & \textcolor{darkred}{69.07 $\pm$ 6.82} & 82.93 $\pm$ 1.82 \\ 
\midrule
\quad + INFONCE & \textcolor{blue}{89.5 $\pm$ 2.00} & \textcolor{blue}{\textbf{84.88 $\pm$ 1.49}} & \textcolor{blue}{91.67 $\pm$ 1.19} \\ 
\quad + SAMI & \textcolor{blue}{91.17 $\pm$ 0.76} & \textcolor{blue}{83.33 $\pm$ 1.07} & \textcolor{blue}{92.86 $\pm$ 1.19} \\
\rowcolor{oursbg}
\quad + \textbf{MIPO} &
\textcolor{blue}{\textbf{93.67 $\pm$ 1.26}}{\scriptsize\color{gain}{+15.7}} &
\textcolor{blue}{80.93 $\pm$ 3.61}{\scriptsize\color{gain}{+8.8}} & 
\textcolor{blue}{\textbf{93.26 $\pm$ 0.69}}{\scriptsize\color{gain}{+13.5}}  \\
\midrule
\textbf{Llama-3.2-3B-Instruct} & 37 & 40.72 & 77.38 \\ 
\rowcolor{basebg}
\quad + Personalized Prompting & 78 & 76.80 & 83.33  \\
\quad + SFT & \textcolor{blue}{84.17 $\pm$ 1.61} & 78.52 $\pm$ 0.79 & \textcolor{blue}{90.48 $\pm$ 1.19}\\
\quad + SFT (Revision) & \textcolor{blue}{81.67 $\pm$ 3.75} & 78.35 $\pm$ 1.86 & \textcolor{blue}{88.89 $\pm$ 0.69} \\
\midrule
\quad + PPO (MI) & \textcolor{darkred}{72.67 $\pm$ 6.05} & \textcolor{darkred}{73.02 $\pm$ 1.66} & \textcolor{blue}{90.08 $\pm$ 3.00} \\
\quad + RLAIF & \textcolor{blue}{90.17 $\pm$ 2.02} & \textcolor{blue}{80.07 $\pm$ 2.84} & \textcolor{blue}{90.48 $\pm$ 1.19}  \\
\quad + RLAIF (MI) & \textcolor{darkred}{74.50 $\pm$ 8.67} & \textcolor{blue}{80.58 $\pm$ 1.19} & \textcolor{blue}{93.65 $\pm$ 2.75}\\
\midrule
\quad + INFONCE & \textcolor{blue}{83.67 $\pm$ 2.75} & 78.69 $\pm$ 0.60 & 
\textcolor{blue}{93.65 $\pm$ 2.75}\\
\quad + SAMI & \textcolor{blue}{\textbf{91.5 $\pm$ 2.29}}  & \textcolor{blue}{82.99 $\pm$ 0.1} & \textbf{\textcolor{blue}{94.84 $\pm$ 1.82}}\\
\rowcolor{oursbg}
\quad + \textbf{MIPO} &
\textcolor{blue}{90.33 $\pm$ 2.08}{\scriptsize\color{gain}{+12.3}}  &
\textbf{\textcolor{blue}{83.41 $\pm$ 0.52}}{\scriptsize\color{gain}{+6.6}}  &
\textbf{\textcolor{blue}{94.84 $\pm$ 1.82}}{\scriptsize\color{gain}{+11.5}} \\
\midrule 
\textbf{Qwen2.5-1.5B-Instruct} & 25.5 & 8.25 & 15.48 \\
\rowcolor{basebg}
\quad + Personalized Prompting & 63.5 & 39.18 & 39.29  \\
\quad + SFT & \textcolor{darkred}{60.5 $\pm$ 0} & \textcolor{darkred}{35.05 $\pm$ 0.52} & 39.68 $\pm$ 3.00  \\
\quad + SFT (Revision) & \textcolor{blue}{67 $\pm$ 2.78} & \textcolor{blue}{46.56 $\pm$ 0.79} & \textcolor{darkred}{35.31 $\pm$ 1.82} \\
\midrule 
\quad + PPO (MI) & \textcolor{darkred}{39.67 $\pm$ 8.81} & \textcolor{darkred}{32.47 $\pm$ 3.61} & 36.90 $\pm$ 2.06\\
\quad + RLAIF & \textcolor{darkred}{50.33 $\pm$ 7.09} & \textcolor{darkred}{13.92 $\pm$ 3.09} & 39.29 $\pm$ 5.46 \\ 
\quad + RLAIF (MI) & \textcolor{darkred}{59.83 $\pm$ 5.84} & \textcolor{darkred}{16.83 $\pm$ 4.79} & \textcolor{darkred}{34.52 $\pm$ 2.38}\\
\midrule 
\quad + INFONCE & \textcolor{darkred}{58.83 $\pm$ 1.04} & \textcolor{blue}{50.69 $\pm$ 2.93} & \textcolor{blue}{59.52 $\pm$ 2.06}\\
\quad + SAMI & \textcolor{blue}{72.83 $\pm$ 1.26} & \textbf{\textcolor{blue}{60.65 $\pm$ 1.58}} & 64.69 $\pm$ 1.82 \\
\rowcolor{oursbg}
\quad + \textbf{MIPO} &
\textbf{\textcolor{blue}{78.83 $\pm$ 2.84}}{\scriptsize\color{gain}{+15.3}}  &
{\textcolor{blue}{60.31 $\pm$ 1.26}}{\scriptsize\color{gain}{+21.1}}  &
\textbf{\textcolor{blue}{74.60 $\pm$ 3.00}}{\scriptsize\color{gain}{+35.31}}  \\
\midrule
\textbf{Qwen2.5-3B-Instruct} & 34 & 11.86 & 46.43 \\
\rowcolor{basebg}
\quad + Personalized Prompting & 76 & 49.49 & 63.10  \\
\quad + SFT & 74.33 $\pm$ 0.29 & \textcolor{darkred}{44.67 $\pm$ 0.79} & 64.29 $\pm$ 3.16  \\
\quad + SFT (Revision) & 74.83 $\pm$ 0.29 & \textcolor{blue}{53.09 $\pm$ 0.52} & 64.68 $\pm$ 2.48  \\
\midrule 
\quad + PPO (MI) & \textcolor{darkred}{72.67 $\pm$ 6.05} & 50.52 $\pm$ 6.08 & \textcolor{blue}{66.67 $\pm$ 1.19}\\
\quad + RLAIF & 77.33 $\pm$ 2.08 & 48.63 $\pm$ 1.30 & \textcolor{blue}{70.64 $\pm$ 2.75}  \\
\quad + RLAIF (MI) & 74.50 $\pm$ 8.67 & 51.72 $\pm$ 3.94 & \textcolor{blue}{72.22 $\pm$ 2.48} \\
\midrule 
\quad + INFONCE & \textcolor{darkred}{70 $\pm$ 1.73} & 50.52 $\pm$ 0.52 & \textcolor{blue}{73.02 $\pm$ 3.00}\\
\quad + SAMI & \textbf{\textcolor{blue}{80.33 $\pm$ 2.57}} & \textcolor{blue}{59.62 $\pm$ 0.30} & \textcolor{blue}{71.83 $\pm$ 4.51}\\
\rowcolor{oursbg}
\quad + \textbf{MIPO} &
\textbf{\textcolor{blue}{80.33 $\pm$ 1.76}}{\scriptsize\color{gain}{+4.3}} &
\textbf{\textcolor{blue}{60.31 $\pm$ 1.36}}{\scriptsize\color{gain}{+10.8}}  &
\textbf{\textcolor{blue}{74.60 $\pm$ 0.69}}{\scriptsize\color{gain}{+11.5}} \\
\midrule
\textbf{Qwen2.5-7B-Instruct} & 39 & 20.62 & 47.62 \\
\rowcolor{basebg}
\quad + Personalized Prompting & 78.5 & 58.25 & 70.24  \\
\quad + SFT & 77 & 56.70 & \textcolor{blue}{75}  \\
\quad + SFT (Revision) & 77 & 59.28 & 72.62  \\
\midrule 
\quad + PPO (MI) & 79 & \textcolor{darkred}{54.12} & 72.62 \\
\quad + RLAIF & 80.5 & \textbf{\textcolor{blue}{63.40}} & \textcolor{blue}{76.19} \\
\quad + RLAIF (MI) & \textbf{\textcolor{blue}{86}} & \textcolor{blue}{64.43} & \textcolor{blue}{\textbf{78.57}} \\
\midrule 
\quad + INFONCE & \textcolor{blue}{82.5} & 57.73 & 72.62 \\
\quad + SAMI & \textcolor{blue}{81.5} & 57.73  & \textcolor{blue}{73.81} \\
\rowcolor{oursbg}
\quad + \textbf{MIPO} &
\textcolor{blue}{81.5}{\scriptsize\color{gain}{+3}} & 58.76 {\scriptsize\color{gain}{+0.5}} & \textcolor{blue}{73.81}{\scriptsize\color{gain}{+3.6}} \\
\bottomrule
\end{tabular}}
% \end{minipage}
\vspace{-0.2cm}
\end{table}

Table \ref{tab:personalization} compares the win rates of MIPO-generated responses to other methods. Since the base models are already instruction-finetuned and receive information about the user-specific context, personalized prompting offers a strong baseline relative to default prompting, which produces generic answers regardless of the user. As a result, personalized prompting (yellow) nearly doubles the win rate over the default responses. MIPO achieves substantial gains on top of personalized prompting: Qwen2.5-7B-Instruct improves by 3\% on average across three domains, the 1--3B models improve by 13--16\%, and Qwen2.5-1.5B-Instruct in particular achieves a 51\% improvement. While RLAIF provides a compelling baseline for some larger models, it performs poorly at the 1B scale. We hypothesize that this is due to unreliable reward signals from smaller critics (see Appendix~\ref{appendix:rm_accuracy}). 

Online mutual information approaches also fall short of MIPO. Both PPO (MI), which rewards mutual information only, and RLAIF (MI), which combines mutual information with the LLM-as-judge reward, underperform across most settings. This aligns with our hypothesis that maximizing the pointwise mutual information under the reference policy serves as a better empirical objective than maximizing the mutual information under the learned policy, since MIPO is less prone to reward hacking. MIPO also outperforms other InfoNCE-based MI implementations across most domains and models, though SAMI also performs comparably.

\begin{table*}[htbp]
\caption{\textbf{Reasoning benchmarks.} GSM and SVAMP use 8-shot; otherwise zero-shot. Reported mean and stddev from 3 seeds. Changes larger than 3 points are highlighted in \textcolor{darkred}{red (+)} and \textcolor{blue}{blue (-)}. Ablations with different negative sampling strategies are reported in Appendix~\ref{appendix:rejection_sampling}.}
\label{tab:rlvr}
\centering
\adjustbox{width=0.8\textwidth}{
\begin{tabular}{lcccccc}
\toprule
Model & GSM (8-shot) {\includegraphics[height=1em]{math_clipart.png}} & SVAMP (8-shot) {\includegraphics[height=1em]{math_clipart.png}}  & ARC-Easy {\includegraphics[height=1em]{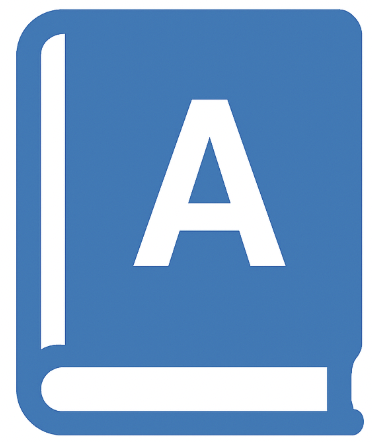}} & ARC-Challenge {\includegraphics[height=1em]{alphabet_clipart.png}} & Avg. \\
\midrule
\rowcolor{basebg}
\textbf{Llama-3.2-1B-Instruct} & 22 & 51.67 & 44 & 33.2 & 37.72 \\
\quad + SFT & 24.83 $\pm$ 1.61{\scriptsize\color{gain}{+2.8}}  & 54.44 $\pm$ 1.50{\scriptsize\color{gain}{+2.8}} & \textcolor{darkred}{34.13 $\pm$ 5.5}{\scriptsize\color{loss}{-9.9}} & \textcolor{darkred}{28.53 $\pm$ 0.64}{\scriptsize\color{loss}{-4.7}} & 35.48{\scriptsize\color{loss}{-2.2}} \\
\rowcolor{oursbg}
\quad + \textbf{MIPO (Ours)} &
\textcolor{blue}{\textbf{29.5 $\pm$ 2.29}}{\scriptsize\color{gain}{+7.5}} &
\textcolor{blue}{\textbf{60.11 $\pm$ 0.19}}{\scriptsize\color{gain}{+8.4}}  &
\textcolor{blue}{51.87 $\pm$ 2.01}{\scriptsize\color{gain}{+7.9}} & 
\textcolor{blue}{39.53 $\pm$ 3.11}{\scriptsize\color{gain}{+6.3}} &  \textcolor{blue}{45.25}{\scriptsize\color{gain}{+7.5}} \\
\quad + SFT (Ground-truth) & 24.17 $\pm$ 1.04{\scriptsize\color{gain}{+2.2}} & 52.56 $\pm$ 0.51{\scriptsize\color{gain}{+0.9}} & \textcolor{darkred}{28.53 $\pm$ 1.10}{\scriptsize\color{loss}{-15.5}} & \textcolor{darkred}{27.13 $\pm$ 2.19}{\scriptsize\color{loss}{-6.1}} & \textcolor{darkred}{33.10}{\scriptsize\color{loss}{-4.6}} \\
\quad + RLAIF (Ground-truth) &
\textcolor{darkred}{10.67 $\pm$ 1.26}{\scriptsize\color{loss}{-11.3}} & 52.56 $\pm$ 2.50{\scriptsize\color{gain}{+0.9}}  & \textcolor{blue}{58.8 $\pm$ 0.92}{\scriptsize\color{gain}{+14.8}} & \textcolor{blue}{42.2 $\pm$ 1.04}{\scriptsize\color{gain}{+9.0}} & \textcolor{blue}{41.06}{\scriptsize\color{gain}{+3.3}} \\
\quad + RLVR (Ground-truth) & 24.83 $\pm$ 3.01{\scriptsize\color{gain}{+2.8}} & \textcolor{blue}{55.89 $\pm$ 3.10}{\scriptsize\color{gain}{+4.2}} & \textcolor{blue}{\textbf{66.2 $\pm$ 2.77}}{\scriptsize\color{gain}{+22.2}}  & \textcolor{blue}{\textbf{43.27 $\pm$ 2.72}}{\scriptsize\color{gain}{+10.1}} & \textbf{\textcolor{blue}{47.55}}{\scriptsize\color{gain}{+9.8}} \\
\midrule

\rowcolor{basebg}
\textbf{Llama-3.2-3B-Instruct} & 71.00 & 78.67 & 80.4 & 68.6 & 74.67 \\
\quad + SFT & \textcolor{darkred}{64.5 $\pm$ 1.32}{\scriptsize\color{loss}{-6.5}}  & \textbf{79.66 $\pm$ 1.53}{\scriptsize\color{gain}{+1.0}} & 77.47 $\pm$ 2.25{\scriptsize\color{loss}{-2.9}} & \textcolor{darkred}{63.67 $\pm$ 0.50}{\scriptsize\color{loss}{-4.9}}  & 71.33 {\scriptsize\color{loss}{-3.3}} \\
\rowcolor{oursbg}
\quad + \textbf{MIPO (Ours)} &
70.17 $\pm$ 2.02{\scriptsize\color{loss}{-0.8}} &
78.22 $\pm$ 1.57{\scriptsize\color{loss}{-0.5}} &
\textcolor{blue}{\textbf{85.26 $\pm$ 0.58}}{\scriptsize\color{gain}{+4.9}} & \textbf{70.93 $\pm$ 0.7}{\scriptsize\color{gain}{+2.3}} &  \textbf{76.15}{\scriptsize\color{gain}{+1.5}} \\
\quad + SFT (Ground-truth) & 69.17 $\pm$ 1.89{\scriptsize\color{loss}{-1.8}} & 77.21 $\pm$ 1.00{\scriptsize\color{loss}{-1.5}} & 78.73 $\pm$ 1.14{\scriptsize\color{loss}{-1.7}} & 70.53 $\pm$ 1.45{\scriptsize\color{gain}{+1.9}} & 73.91{\scriptsize\color{loss}{-0.8}} \\
\quad + RLAIF (Ground-truth) & \textbf{71.67 $\pm$ 1.76}{\scriptsize\color{gain}{+0.7}} & 79.11 $\pm$ 2.99{\scriptsize\color{gain}{+0.4}} & 80.07 $\pm$ 2.73{\scriptsize\color{loss}{-0.3}} & 70.33 $\pm$ 1.80{\scriptsize\color{gain}{+1.7}} & 75.30{\scriptsize\color{gain}{+0.6}} \\
\quad + RLVR (Ground-truth) & \textcolor{darkred}{65.17 $\pm$ 3.55}{\scriptsize\color{loss}{-5.8}} & 79.44 $\pm$ 0.51{\scriptsize\color{gain}{+0.8}} & 83.20 $\pm$ 1.40{\scriptsize\color{gain}{+2.8}} & 69.60 $\pm$ 1.20{\scriptsize\color{gain}{+1}} & 74.35{\scriptsize\color{loss}{-0.3}}\\

\midrule
\rowcolor{basebg}
\textbf{Qwen2.5-1.5B-Instruct} & 65.5 & 82.33 & 79 & 63.6 & 72.61  \\
\quad + SFT & 67 $\pm$ 3.50{\scriptsize\color{gain}{+1.5}}  & 82.00 $\pm$ 1.53{\scriptsize\color{loss}{-0.3}} & \textcolor{darkred}{66.07 $\pm$ 1.51}{\scriptsize\color{loss}{-12.9}} & \textcolor{darkred}{52.47 $\pm$ 1.03}{\scriptsize\color{loss}{-11.1}} & \textcolor{darkred}{66.89}{\scriptsize\color{loss}{}} \\

\rowcolor{oursbg}
\quad + \textbf{MIPO (Ours)} &
\textcolor{blue}{\textbf{71 $\pm$ 1.80}}{\scriptsize\color{gain}{+5.5}} &
81.67 $\pm$ 2.03{\scriptsize\color{loss}{-0.7}} &
\textcolor{blue}{82.27 $\pm$ 0.81}{\scriptsize\color{gain}{+3.3}} & 65.93 $\pm$ 1.70{\scriptsize\color{gain}{+2.3}} &  75.22{\scriptsize\color{gain}{+2.4}} \\

\quad + SFT (Ground-truth) & 67.50 $\pm$ 2.00{\scriptsize\color{gain}{+2.0}} & 81.33 $\pm$ 0.34{\scriptsize\color{loss}{-1.0}}  & \textcolor{darkred}{75.2 $\pm$ 0.87}{\scriptsize\color{loss}{-3.8}}  & 62.2 $\pm$ 1.83{\scriptsize\color{loss}{-1.4}}  & 71.56{\scriptsize\color{loss}{-1.3}} \\

\quad + RLAIF (Ground-truth) & \textcolor{darkred}{62.5 $\pm$ 7.76}{\scriptsize\color{loss}{-3.0}} & \textbf{83 $\pm$ 1.67}{\scriptsize\color{gain}{+0.7}} & \textcolor{blue}{85.4 $\pm$ 0.2}{\scriptsize\color{gain}{+6.4}} & \textcolor{blue}{\textbf{70.47 $\pm$ 0.76}}{\scriptsize\color{gain}{+6.9}} & \textcolor{blue}{\textbf{75.34}}{\scriptsize\color{gain}{+2.5}} \\

\quad + RLVR (Ground-truth) & \textcolor{darkred}{62.33 $\pm$ 5.97}{\scriptsize\color{loss}{-3.2}} & 79.78 $\pm$ 0.96{\scriptsize\color{loss}{-2.6}} & \textcolor{blue}{\textbf{86.73 $\pm$ 0.70}}{\scriptsize\color{gain}{+4.4}} &  69.13 $\pm$ 3.59{\scriptsize\color{gain}{+5.5}} & 74.49{\scriptsize\color{gain}{+1.6}}\\
\midrule

\rowcolor{basebg}
\textbf{Qwen2.5-3B-Instruct} & 84.5 & 90.67 & 92 & 79.4 & 86.64 \\
\quad + SFT & 85.83 $\pm$ 0.76{\scriptsize\color{gain}{+1.3}}  & 89.78 $\pm$ 1.02{\scriptsize\color{loss}{-0.9}}  & \textcolor{darkred}{87.47 $\pm$ 0.23}{\scriptsize\color{loss}{-4.5}} & 78.67 $\pm$ 1.17{\scriptsize\color{loss}{-0.7}} & 85.44{\scriptsize\color{loss}{-1.2}} \\
\rowcolor{oursbg}
\quad + \textbf{MIPO (Ours)} &
\textcolor{blue}{\textbf{89.17 $\pm$ 3.75}}{\scriptsize\color{gain}{+4.7}} &
\textbf{91.33 $\pm$ 1.73}{\scriptsize\color{gain}{+0.7}}  &
90.80 $\pm$ 0.35{\scriptsize\color{loss}{-1.2}} & \textbf{80.13 $\pm$ 2.01}{\scriptsize\color{gain}{+0.7}} &  \textbf{87.86}{\scriptsize\color{gain}{+1.2}} \\
\quad + SFT (Ground-truth) & 83.83 $\pm$ 2.47{\scriptsize\color{loss}{-0.7}} & 90.56 $\pm$ 0.20{\scriptsize\color{loss}{-0.1}}  & 90.33 $\pm$ 0.99{\scriptsize\color{loss}{-1.7}} & 79.6 $\pm$ 1.11{\scriptsize\color{gain}{+0.2}} & 86.08{\scriptsize\color{loss}{-0.5}} \\
\quad + RLAIF (Ground-truth) & 82.83 $\pm$ 1.53{\scriptsize\color{loss}{-1.7}} & 90.22 $\pm$ 0.69{\scriptsize\color{loss}{-0.5}} & \textbf{91.8 $\pm$ 0.2}{\scriptsize\color{loss}{-0.2}} & 79.4 $\pm$ 1.25 & 86.06{\scriptsize\color{loss}{-0.6}} \\
\quad + RLVR (Ground-truth) & \textcolor{darkred}{69.83 $\pm$ 4.31}{\scriptsize\color{loss}{-14.7}} & \textcolor{darkred}{85.89 $\pm$ 4.40}{\scriptsize\color{loss}{-4.8}} & 91.47 $\pm$ 0.46{\scriptsize\color{loss}{-0.5}} & 78.60 $\pm$ 0.60{\scriptsize\color{loss}{-0.8}} & 81.45{\scriptsize\color{loss}{-5.2}}\\
\midrule
\rowcolor{basebg}
\textbf{Qwen2.5-7B-Instruct} & 93.5 & 91.33 & 93.80 & 88.20 & 91.71 \\
\quad + SFT & \textbf{92.50}{\scriptsize\color{loss}{-1.0}} & 92.00{\scriptsize\color{gain}{+0.7}} & 93.60{\scriptsize\color{loss}{-0.2}} & 87.80{\scriptsize\color{loss}{-0.4}} & 91.48{\scriptsize\color{loss}{-0.2}} \\

\rowcolor{oursbg}
\quad + \textbf{MIPO (Ours)} & \textbf{93.00}{\scriptsize\color{loss}{-0.5}} & 91.67{\scriptsize\color{gain}{+0.3}} & 93.80 & \textbf{90.4}{\scriptsize\color{gain}{+2.2}} & \textbf{92.22}{\scriptsize\color{gain}{+0.5}} \\ 

\quad + SFT (Ground-truth) & 92.00{\scriptsize\color{loss}{-1.5}} & 91.33  & 93.60{\scriptsize\color{loss}{-0.2}} & 90.00{\scriptsize\color{gain}{+1.8}} & 91.73 \\ 

\quad + RLAIF (Ground-truth) & 92.50{\scriptsize\color{loss}{-1.0}} & 91.33 & \textbf{94.20{\scriptsize\color{gain}{+0.4}}} & 88.60{\scriptsize\color{gain}{+0.4}} & 91.66{\scriptsize\color{loss}{-0.1}} \\ 

\quad + RLVR (Ground-truth) & \textcolor{darkred}{90.5}{\scriptsize\color{loss}{-3}} & 90.67{\scriptsize\color{loss}{-0.7}}  & \textbf{94.20{\scriptsize\color{gain}{+0.4}}} & 86.4{\scriptsize\color{loss}{-1.8}} & 90.44{\scriptsize\color{loss}{-1.3}} \\ 
\bottomrule
\vspace{-1.2cm}
\end{tabular}}
\end{table*}

Examples in Appendix~\ref{appendix:personalization_examples} show qualitative improvements in MIPO responses. For example, the prompt presents a conversation between two colleagues discussing lunch options: one speaker says they'll join their co-worker at a coffee shop to get a salad (\emph{``No, I think I'll come with you. I'm longing for a nice salad."}). The model is asked to discuss the speaker's motivation, with specific instructions to consider factors such as social dynamics between colleagues and emotions driving the speaker's response. Personalized prompted responses emphasize the speaker's desire for a salad rather than social politeness or companionship. In contrast, MIPO suggests motivations grounded in the specific aspects the user asked about.

\begin{figure}[h]
\begin{center}
\centerline{\includegraphics[width=\columnwidth]{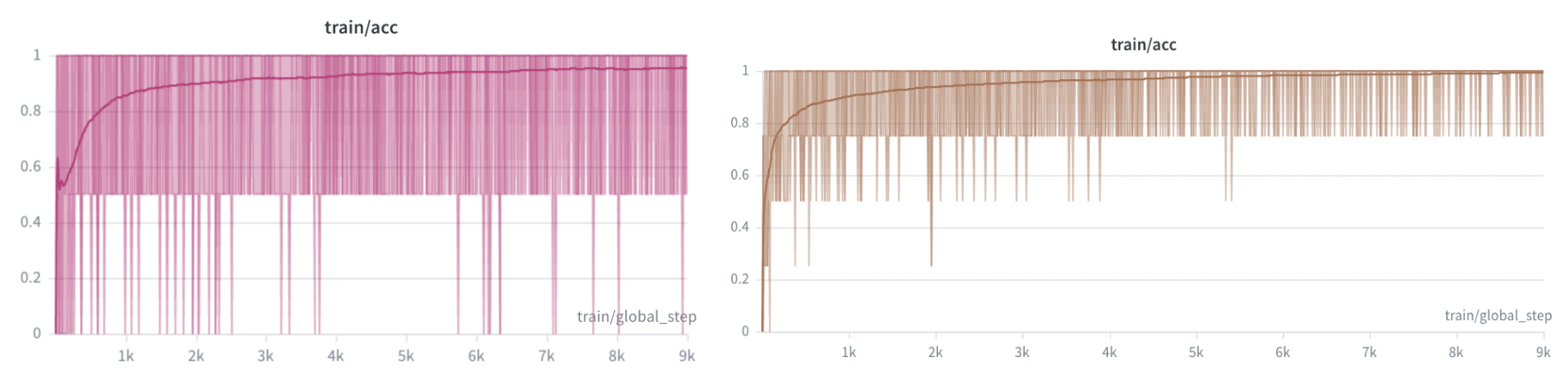}}
\caption{\textbf{Learning curves (positive and negative prediction accuracy) during training.} Left: Qwen2.5-3B-Instruct; right: Llama3-1B-Instruct on Multifaceted Bench with $N=10$ negatives per positive. The learning curves eventually plateau around 6–7k training steps, suggesting that the gains of self-improvement eventually saturate.}
\label{fig:learning_curves}
\end{center}
\vspace{-0.5cm}
\end{figure}

\textbf{R2: MIPO improves models' output diversity via personalization.} Maximizing conditional MI increases diversity because variation in $c$ produces different outputs $y$ even for the same prompt. Self-BLEU-4 scores~\citep{zhu2018texygenbenchmarkingplatformtext} in Table \ref{tab:diversity} confirm this empirically. MIPO lowers average self-BLEU for all models. In contrast, SFT yields higher self-BLEU on average (see Table~\ref{tab:self_bleu_diversity} for individual task-model pair scores). This result is especially meaningful given recent observations of LLM homogeneity~\citep{jiang2025artificialhivemindopenendedhomogeneity, zhang2025verbalizedsamplingmitigatemode, kirk2024understandingeffectsrlhfllm, abdulhai2026llmsdistortwrittenlanguage}, since it suggests that personalization and in-context steerability could help mitigate mode collapse. 
% 
% While diversity is not directly targeted as a reward, using Mutual Information reward improves output diversity by guiding the models to personalize to each user's unique contexts.
\begin{table}[h]
% \vspace{-0.1cm}
% \centering
% \begin{minipage}{0.5\textwidth}
\caption{\textbf{Self-BLEU-4~\cite{zhu2018texygenbenchmarkingplatformtext} from pre- and post-training averaged across three personalization tasks.} Lower values indicate greater diversity (\textcolor{darkgreen}{\ding{51}} indicates diversity improvement). See Table~\ref{tab:self_bleu_diversity} for individual task-model scores.} 
\label{tab:diversity}
\centering
\adjustbox{width=0.8\columnwidth}{
\begin{tabular}{lccccc}
\toprule
Model & Pre-Training & Post-SFT & Post-MIPO  \\
\midrule
Llama 1B & 0.420 & 0.427 \textcolor{red}{\ding{55}} & \textbf{0.393}  \textcolor{darkgreen}{\ding{51}} \\
Llama 3B & 0.379 & 0.389 \textcolor{red}{\ding{55}} & \textbf{0.371} \textcolor{darkgreen}{\ding{51}} \\
Qwen 1.5B & 0.310 & 0.316 \textcolor{red}{\ding{55}} & \textbf{0.256}  \textcolor{darkgreen}{\ding{51}} \\
Qwen 3B & 0.312 & 0.316 \textcolor{red}{\ding{55}} & \textbf{0.272} \textcolor{darkgreen}{\ding{51}} \\
Qwen 7B & 0.310 & 0.318 \textcolor{red}{\ding{55}} & \textbf{0.309} \textcolor{darkgreen}{\ding{51}} \\
\bottomrule
\end{tabular}}
\vspace{-0.2cm}
\end{table}
\sethlcolor{yellow!40}

\begin{figure}[h]
\begin{center}
\centerline{\includegraphics[width=\columnwidth]{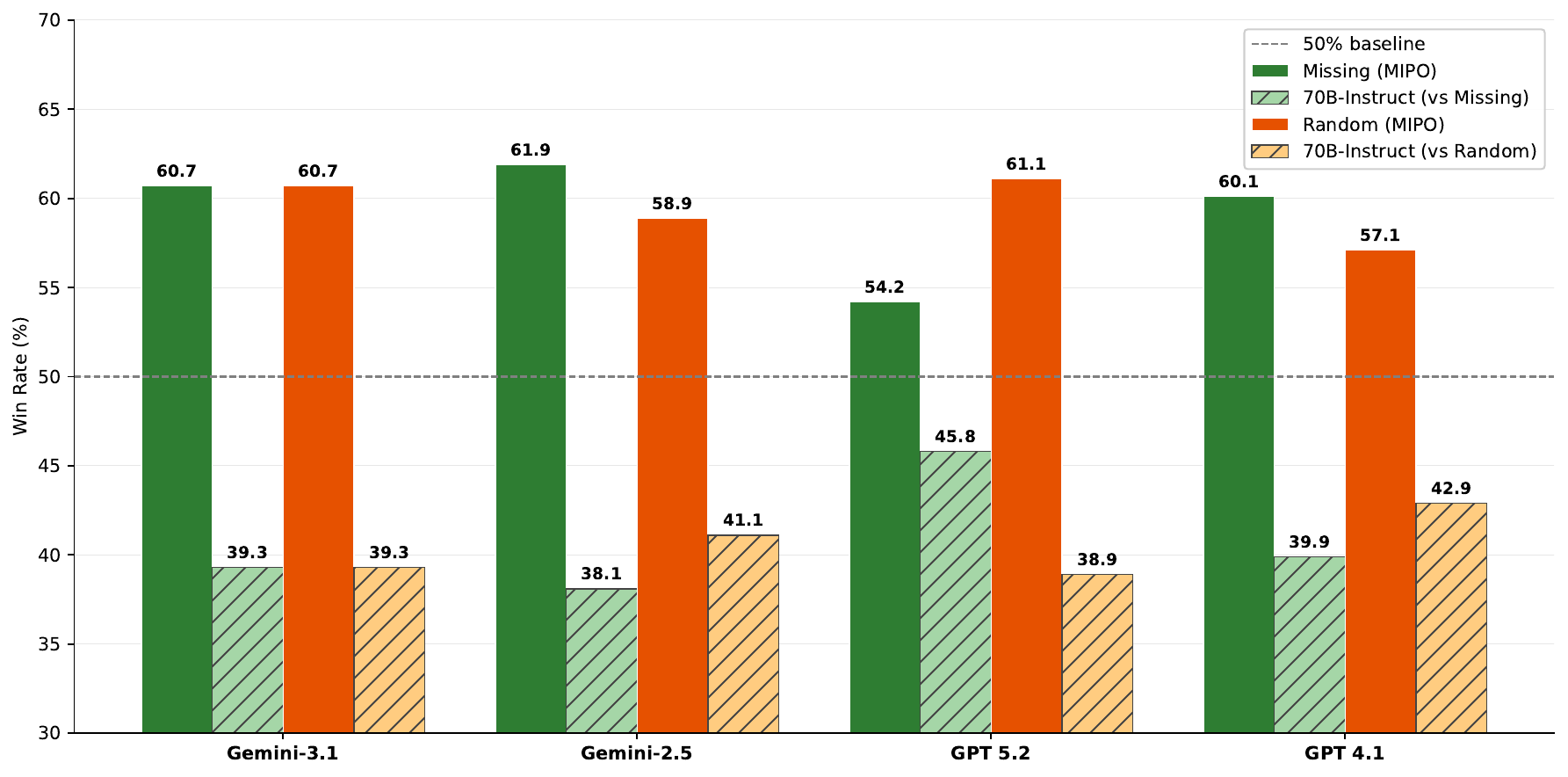}}
\caption{\textbf{Win rates of MIPO versus instruct-tuned models.} We report the win rates of MIPO against the baseline (y-axis) across a suite of LLM-judges (x-axis).}
\label{fig:win_rate_70b}
\end{center}
\vspace{-0.8cm}
\end{figure}

\textbf{R3: MIPO gains scale to 70B models, achieving 54--62\% win rates against personalized prompting baselines.} We train Llama-3.3-70B-Instruct with two versions of MIPO (using missing and random contexts as negatives) on Multifaceted Bench using LoRA rank 32~\citep{hu2021loralowrankadaptationlarge}, and compare both versions against the personalized prompting baseline. For evaluation, we present each pair of responses to four LLM judges (GPT-4.1, GPT-5.2, Gemini 2.5 Flash, and Gemini 3.1 Flash Lite Preview). Each judge selects the preferred response according to a personalization rubric based on the presented user context (see Appendix~\ref{appendix:binary_evaluation} for the prompts). Fig.~\ref{fig:win_rate_70b} shows that MIPO's benefits are consistently observed at the 70B scale across all four LLM judges. 

So far, our empirical results have established MIPO as an effective personalization method that yields improvements without any human supervision or additional data. However, does MIPO generalize to tasks beyond personalization? Unlike prior work on constitution following~\citep{fränken2024selfsupervisedalignmentmutualinformation}, MIPO extends to more general settings where prompts and contexts are not decoupled. When only prompts are given, chosen responses are paired with negatives generated from \emph{random} prompts $x' \not= x$. We evaluate this approach on a range of LLM benchmarks, including math and MCQ. While we do not expect MIPO to outperform RLVR, since RLVR requires access to ground-truth solutions and verifiers, we test whether MIPO can serve as a meaningful alternative despite having no external feedback.

\textbf{R4: Surprisingly, MIPO generalizes to math and MCQ tasks, achieving 1--4\% improvements overall and an 20\% improvement for smaller 1B Llama models.} In Table~\ref{tab:rlvr}, the benefit of the contrastive signal is highlighted by the gap between SFT (without ground-truth) and MIPO. SFT performs poorly, likely because the smaller base models' initial outputs are of low quality. By contrast, MIPO's contrastive pairs yield substantial improvements over the instruction-following baselines, even at small model scales. MIPO also occasionally outperforms RLVR and RLAIF, both of which have access to ground-truth answers during training. MIPO shows impressive gains especially for smaller 1B models with raw point increases of 2.4--7.5 on average across tasks. Although performance gains vary by domain and model, we hypothesize that this reflects how heavily a particular model has been fine-tuned for specific tasks. For example, we suspect Qwen2.5-7B-Instruct has been heavily fine-tuned on math benchmarks like GSM, leaving little room for further improvement.

\textbf{R5: We investigate the effects of different negative sampling strategies} (Appendix Table~\ref{appendix:rejection_sampling}). For personalized instruction-following, sampling rejected responses with random or missing contexts substantially outperforms reshuffling: random sampling outperforms reshuffling by 1.3--13\% (varying across model and domain), and missing-context sampling outperforms it by 3--30\%. This is likely because the conditional MI objective requires the chosen and rejected responses to differ only in user context, not in the prompt. Reshuffling violates this assumption by pairing responses with different prompts as well as different contexts. For math and MCQ benchmarks, by contrast, reshuffling performs comparably to random sampling. Although re-using chosen responses violates the independence assumption between positives and negatives, this does not appear to hurt empirical performance on these benchmarks.

To test whether negatives need to come from the reference policy, we conduct ablations by pairing chosen responses with random alphanumeric strings of the same length as the rejected responses (Appendix Table ~\ref{appendix:n_sampling}). Although these negatives still provide a clear contrastive signal against the positives, they fail to match MIPO's gains. Table ~\ref{appendix:n_sampling} also shows that increasing the number of negatives per positive further improves MIPO's performance, which is consistent with the theoretical understanding that larger $N$ tightens the lower bound on mutual information.

\section{Conclusion \& Discussion}
We propose MIPO, a contrastive data augmentation method based on maximizing mutual information between prompts and responses, and evaluate it across a suite of both verifiable and non-verifiable tasks. Our empirical evaluation shows that: 
\begin{itemize}[noitemsep, topsep=0pt]
\item MIPO provides effective personalization, improving models by 3--51\% across three pluralistic user datasets;
\item MIPO also extends to a broader set of LLM tasks, including math and MCQ, achieving an additional 1--4\% improvement on top of already instruction-finetuned models; 
\item MIPO yields especially large gains for smaller models whose self-generated data is likely suboptimal (e.g., 20\% average improvement for Llama3.2-1B-Instruct).
\end{itemize} Crucially, MIPO can achieve these gains \emph{without any additional data or human supervision} and requires only a simple data-processing step before standard DPO training. These results suggest a promising direction for self-improvement using LLMs' implicit reward signals.

\textbf{Limitations.} \textbf{(i) Reliability of LLM judges.} Our current personalization results rely on LLM-as-judge for evaluation, which has both advantages and known limitations. While LLM-based evaluation is often used for cost-effectiveness, replicability, and high accuracy~\footnote{For example, AlpacaEval 2.0~\citep{dubois2025lengthcontrolledalpacaevalsimpleway} uses GPT-4 as a judge and demonstrates strong correlation with human evaluations from Chatbot Arena (Spearman correlation of 0.93--0.98 based on 20K annotations).}, it can also exhibit biases, such as the model's preference for verbosity. Therefore, when using LLM judges, it is crucial to validate them against human preference judgments or ground-truth data. \textbf{(ii) Using GPT-4o to obtain user contexts from real-world data.} For personalization, we used GPT-4o as a preprocessing step to obtain user contexts, which are not directly available in many real-user datasets (e.g., PRISM and Community Alignment). This information is made available to all baseline methods, and we evaluate only the policy’s ability to adapt to these user-specific details in its response. An alternative to using GPT-4o would be to use past conversations or self-stated preferences. However, self-stated preferences are typically unavailable in real-world datasets unless explicitly collected via surveys~\citep{kirk2024prismalignmentdatasetparticipatory}. Moreover, even state-of-the-art reward models conditioned on self-stated preferences or past conversations achieve only 60-62\% accuracy in preference prediction~\citep{nam2026learningsummarizeuserinformation}, which limits their reliability for judging personalized response preferability. We consider this a crucial limitation of many existing personalization datasets and see this as an important area of future work. \textbf{(iii) Noisy training signals.} Since MIPO's training signal is derived from the model itself, its effectiveness depends on the model's initial capabilities and the diversity of training prompts. When prompts or contexts are too similar to each other, the contrastive signal may be too weak to drive learning effectively.

% Furthermore, these results yield important insights for future direction: the effectiveness of negative samples for model improvement can vary, so identifying and generating informative rejections can be a crucial next step to further the gains of MIPO
% - We evaluate MIPO on both verifiable and non-verifiable tasks; on verifiable tasks, we observe moderate improvements across all model and task pairs, but crucially show that smaller models can learn from suboptimal self-generated data (which other methods without privileged information cannot achieve). 

% - Finding interesting tasks that can push the model's performance boundaries is challenging for any post-training problems (still within pre-training distribution but challenging for the curent model capacities); and we find personalization tasks particularly interesting because they are novel (not in training data) but within model's capabilities (esp. for bigger models)

% - On the other hand, we see significant improvements on non-verifiable tasks (these are also harder, difficult to improve \& evaluate tasks in real life); we use personalization as a representative task in this category (as demonstration / testing bed for the model's \emph{in-context steerability}).

\clearpage
\section*{Impact Statement} Homogenization of model outputs is increasingly becoming an important AI safety concern~\citep{abdulhai2026llmsdistortwrittenlanguage}. We believe that more effective steering of model responses and improved personalization can help mitigate this issue. We propose a self-training method that maximizes the mutual information between prompts and model responses, and, in the personalization setting, the conditional mutual information between user contexts and model responses conditioned on the prompt. Our empirical results are promising in showing that models can self-improve without external feedback or additional data. However, this does not downplay the role of human oversight in model development. Human (preference) data remains critical but is often expensive to collect. We show that an implicit reward signal based on mutual information can serve as a useful alternative when such data is difficult to gather or scale. This may allow human data collection efforts to focus instead on model evaluation and deployment oversight, where they are most needed to ensure safe and aligned model behavior, and reduce the burden of human data in model training cycles.

\section*{Acknowledgment} This research was supported by the UW-Amazon Science Gift Hub, UW-Tsukuba Amazon NVIDIA Cross Pacific AI Initiative (XPAI), Sony Research Award, Tinker Research Grants, Character.AI, DoorDash, Open Philanthropy, Coefficient Giving, Toyota Research Institute, and the Schmidt AI2050 Fellows program. This material is based upon work supported by the Defense Advanced Research Projects Agency and the Air Force Research Laboratory, contract number(s): FA8650-23-C-7316. Any opinions, findings and conclusions, or recommendations expressed in this material are those of the author(s) and do not necessarily reflect the views of AFRL or DARPA.

\bibliographystyle{icml2026}
\bibliography{example_paper}

@misc{bai2022constitutionalaiharmlessnessai,
      title={Constitutional AI: Harmlessness from AI Feedback}, 
      author={Yuntao Bai and Saurav Kadavath and Sandipan Kundu and Amanda Askell and Jackson Kernion and Andy Jones and Anna Chen and Anna Goldie and Azalia Mirhoseini and Cameron McKinnon and Carol Chen and Catherine Olsson and Christopher Olah and Danny Hernandez and Dawn Drain and Deep Ganguli and Dustin Li and Eli Tran-Johnson and Ethan Perez and Jamie Kerr and Jared Mueller and Jeffrey Ladish and Joshua Landau and Kamal Ndousse and Kamile Lukosuite and Liane Lovitt and Michael Sellitto and Nelson Elhage and Nicholas Schiefer and Noemi Mercado and Nova DasSarma and Robert Lasenby and Robin Larson and Sam Ringer and Scott Johnston and Shauna Kravec and Sheer El Showk and Stanislav Fort and Tamera Lanham and Timothy Telleen-Lawton and Tom Conerly and Tom Henighan and Tristan Hume and Samuel R. Bowman and Zac Hatfield-Dodds and Ben Mann and Dario Amodei and Nicholas Joseph and Sam McCandlish and Tom Brown and Jared Kaplan},
      year={2022},
      eprint={2212.08073},
      archivePrefix={arXiv},
      primaryClass={cs.CL},
      url={https://arxiv.org/abs/2212.08073}, 
}

@article{jaques2019way,
  title={Way off-policy batch deep reinforcement learning of implicit human preferences in dialog},
  author={Jaques, Natasha and Ghandeharioun, Asma and Shen, Judy Hanwen and Ferguson, Craig and Lapedriza, Agata and Jones, Noah and Gu, Shixiang and Picard, Rosalind},
  journal={arXiv preprint arXiv:1907.00456},
  year={2019}
}

@inproceedings{jaques2017sequence,
  title={Sequence tutor: Conservative fine-tuning of sequence generation models with kl-control},
  author={Jaques, Natasha and Gu, Shixiang and Bahdanau, Dzmitry and Hern{\'a}ndez-Lobato, Jos{\'e} Miguel and Turner, Richard E and Eck, Douglas},
  booktitle={International Conference on Machine Learning},
  pages={1645--1654},
  year={2017},
  organization={PMLR}
}

@misc{lee2024aligningthousandspreferencesmessage,
      title={Aligning to Thousands of Preferences via System Message Generalization}, 
      author={Seongyun Lee and Sue Hyun Park and Seungone Kim and Minjoon Seo},
      year={2024},
      eprint={2405.17977},
      archivePrefix={arXiv},
      primaryClass={cs.CL},
      url={https://arxiv.org/abs/2405.17977}, 
}

@misc{jiang2025artificialhivemindopenendedhomogeneity,
      title={Artificial Hivemind: The Open-Ended Homogeneity of Language Models (and Beyond)}, 
      author={Liwei Jiang and Yuanjun Chai and Margaret Li and Mickel Liu and Raymond Fok and Nouha Dziri and Yulia Tsvetkov and Maarten Sap and Alon Albalak and Yejin Choi},
      year={2025},
      eprint={2510.22954},
      archivePrefix={arXiv},
      primaryClass={cs.CL},
      url={https://arxiv.org/abs/2510.22954}, 
}

@misc{kirk2024understandingeffectsrlhfllm,
      title={Understanding the Effects of RLHF on LLM Generalisation and Diversity}, 
      author={Robert Kirk and Ishita Mediratta and Christoforos Nalmpantis and Jelena Luketina and Eric Hambro and Edward Grefenstette and Roberta Raileanu},
      year={2024},
      eprint={2310.06452},
      archivePrefix={arXiv},
      primaryClass={cs.LG},
      url={https://arxiv.org/abs/2310.06452}, 
}

@misc{li2024personalizedlanguagemodelingpersonalized,
      title={Personalized Language Modeling from Personalized Human Feedback}, 
      author={Xinyu Li and Ruiyang Zhou and Zachary C. Lipton and Liu Leqi},
      year={2024},
      eprint={2402.05133},
      archivePrefix={arXiv},
      primaryClass={cs.CL},
      url={https://arxiv.org/abs/2402.05133}, 
}

@misc{yuan2025selfrewardinglanguagemodels,
      title={Self-Rewarding Language Models}, 
      author={Weizhe Yuan and Richard Yuanzhe Pang and Kyunghyun Cho and Xian Li and Sainbayar Sukhbaatar and Jing Xu and Jason Weston},
      year={2025},
      eprint={2401.10020},
      archivePrefix={arXiv},
      primaryClass={cs.CL},
      url={https://arxiv.org/abs/2401.10020}, 
}

@article{Guo_2025,
   title={DeepSeek-R1 incentivizes reasoning in LLMs through reinforcement learning},
   volume={645},
   ISSN={1476-4687},
   url={http://dx.doi.org/10.1038/s41586-025-09422-z},
   DOI={10.1038/s41586-025-09422-z},
   number={8081},
   journal={Nature},
   publisher={Springer Science and Business Media LLC},
   author={Guo, Daya and Yang, Dejian and Zhang, Haowei and Song, Junxiao and Wang, Peiyi and Zhu, Qihao and Xu, Runxin and Zhang, Ruoyu and Ma, Shirong and Bi, Xiao and Zhang, Xiaokang and Yu, Xingkai and Wu, Yu and Wu, Z. F. and Gou, Zhibin and Shao, Zhihong and Li, Zhuoshu and Gao, Ziyi and Liu, Aixin and Xue, Bing and Wang, Bingxuan and Wu, Bochao and Feng, Bei and Lu, Chengda and Zhao, Chenggang and Deng, Chengqi and Ruan, Chong and Dai, Damai and Chen, Deli and Ji, Dongjie and Li, Erhang and Lin, Fangyun and Dai, Fucong and Luo, Fuli and Hao, Guangbo and Chen, Guanting and Li, Guowei and Zhang, H. and Xu, Hanwei and Ding, Honghui and Gao, Huazuo and Qu, Hui and Li, Hui and Guo, Jianzhong and Li, Jiashi and Chen, Jingchang and Yuan, Jingyang and Tu, Jinhao and Qiu, Junjie and Li, Junlong and Cai, J. L. and Ni, Jiaqi and Liang, Jian and Chen, Jin and Dong, Kai and Hu, Kai and You, Kaichao and Gao, Kaige and Guan, Kang and Huang, Kexin and Yu, Kuai and Wang, Lean and Zhang, Lecong and Zhao, Liang and Wang, Litong and Zhang, Liyue and Xu, Lei and Xia, Leyi and Zhang, Mingchuan and Zhang, Minghua and Tang, Minghui and Zhou, Mingxu and Li, Meng and Wang, Miaojun and Li, Mingming and Tian, Ning and Huang, Panpan and Zhang, Peng and Wang, Qiancheng and Chen, Qinyu and Du, Qiushi and Ge, Ruiqi and Zhang, Ruisong and Pan, Ruizhe and Wang, Runji and Chen, R. J. and Jin, R. L. and Chen, Ruyi and Lu, Shanghao and Zhou, Shangyan and Chen, Shanhuang and Ye, Shengfeng and Wang, Shiyu and Yu, Shuiping and Zhou, Shunfeng and Pan, Shuting and Li, S. S. and Zhou, Shuang and Wu, Shaoqing and Yun, Tao and Pei, Tian and Sun, Tianyu and Wang, T. and Zeng, Wangding and Liu, Wen and Liang, Wenfeng and Gao, Wenjun and Yu, Wenqin and Zhang, Wentao and Xiao, W. L. and An, Wei and Liu, Xiaodong and Wang, Xiaohan and Chen, Xiaokang and Nie, Xiaotao and Cheng, Xin and Liu, Xin and Xie, Xin and Liu, Xingchao and Yang, Xinyu and Li, Xinyuan and Su, Xuecheng and Lin, Xuheng and Li, X. Q. and Jin, Xiangyue and Shen, Xiaojin and Chen, Xiaosha and Sun, Xiaowen and Wang, Xiaoxiang and Song, Xinnan and Zhou, Xinyi and Wang, Xianzu and Shan, Xinxia and Li, Y. K. and Wang, Y. Q. and Wei, Y. X. and Zhang, Yang and Xu, Yanhong and Li, Yao and Zhao, Yao and Sun, Yaofeng and Wang, Yaohui and Yu, Yi and Zhang, Yichao and Shi, Yifan and Xiong, Yiliang and He, Ying and Piao, Yishi and Wang, Yisong and Tan, Yixuan and Ma, Yiyang and Liu, Yiyuan and Guo, Yongqiang and Ou, Yuan and Wang, Yuduan and Gong, Yue and Zou, Yuheng and He, Yujia and Xiong, Yunfan and Luo, Yuxiang and You, Yuxiang and Liu, Yuxuan and Zhou, Yuyang and Zhu, Y. X. and Huang, Yanping and Li, Yaohui and Zheng, Yi and Zhu, Yuchen and Ma, Yunxian and Tang, Ying and Zha, Yukun and Yan, Yuting and Ren, Z. Z. and Ren, Zehui and Sha, Zhangli and Fu, Zhe and Xu, Zhean and Xie, Zhenda and Zhang, Zhengyan and Hao, Zhewen and Ma, Zhicheng and Yan, Zhigang and Wu, Zhiyu and Gu, Zihui and Zhu, Zijia and Liu, Zijun and Li, Zilin and Xie, Ziwei and Song, Ziyang and Pan, Zizheng and Huang, Zhen and Xu, Zhipeng and Zhang, Zhongyu and Zhang, Zhen},
   year={2025},
   month=sep, pages={633–638} }

@misc{zhang2025verbalizedsamplingmitigatemode,
      title={Verbalized Sampling: How to Mitigate Mode Collapse and Unlock LLM Diversity}, 
      author={Jiayi Zhang and Simon Yu and Derek Chong and Anthony Sicilia and Michael R. Tomz and Christopher D. Manning and Weiyan Shi},
      year={2025},
      eprint={2510.01171},
      archivePrefix={arXiv},
      primaryClass={cs.CL},
      url={https://arxiv.org/abs/2510.01171}, 
}

@misc{sorensen2025spectrumtuningposttrainingdistributional,
      title={Spectrum Tuning: Post-Training for Distributional Coverage and In-Context Steerability}, 
      author={Taylor Sorensen and Benjamin Newman and Jared Moore and Chan Park and Jillian Fisher and Niloofar Mireshghallah and Liwei Jiang and Yejin Choi},
      year={2025},
      eprint={2510.06084},
      archivePrefix={arXiv},
      primaryClass={cs.CL},
      url={https://arxiv.org/abs/2510.06084}, 
}

@misc{bai2022traininghelpfulharmlessassistant,
      title={Training a Helpful and Harmless Assistant with Reinforcement Learning from Human Feedback}, 
      author={Yuntao Bai and Andy Jones and Kamal Ndousse and Amanda Askell and Anna Chen and Nova DasSarma and Dawn Drain and Stanislav Fort and Deep Ganguli and Tom Henighan and Nicholas Joseph and Saurav Kadavath and Jackson Kernion and Tom Conerly and Sheer El-Showk and Nelson Elhage and Zac Hatfield-Dodds and Danny Hernandez and Tristan Hume and Scott Johnston and Shauna Kravec and Liane Lovitt and Neel Nanda and Catherine Olsson and Dario Amodei and Tom Brown and Jack Clark and Sam McCandlish and Chris Olah and Ben Mann and Jared Kaplan},
      year={2022},
      eprint={2204.05862},
      archivePrefix={arXiv},
      primaryClass={cs.CL},
      url={https://arxiv.org/abs/2204.05862}, 
}

@misc{kirk2024prismalignmentdatasetparticipatory,
      title={The PRISM Alignment Dataset: What Participatory, Representative and Individualised Human Feedback Reveals About the Subjective and Multicultural Alignment of Large Language Models}, 
      author={Hannah Rose Kirk and Alexander Whitefield and Paul Röttger and Andrew Bean and Katerina Margatina and Juan Ciro and Rafael Mosquera and Max Bartolo and Adina Williams and He He and Bertie Vidgen and Scott A. Hale},
      year={2024},
      eprint={2404.16019},
      archivePrefix={arXiv},
      primaryClass={cs.CL},
      url={https://arxiv.org/abs/2404.16019}, 
}

@misc{fränken2024selfsupervisedalignmentmutualinformation,
      title={Self-Supervised Alignment with Mutual Information: Learning to Follow Principles without Preference Labels}, 
      author={Jan-Philipp Fränken and Eric Zelikman and Rafael Rafailov and Kanishk Gandhi and Tobias Gerstenberg and Noah D. Goodman},
      year={2024},
      eprint={2404.14313},
      archivePrefix={arXiv},
      primaryClass={cs.CL},
      url={https://arxiv.org/abs/2404.14313}, 
}

@misc{zhu2018texygenbenchmarkingplatformtext,
      title={Texygen: A Benchmarking Platform for Text Generation Models}, 
      author={Yaoming Zhu and Sidi Lu and Lei Zheng and Jiaxian Guo and Weinan Zhang and Jun Wang and Yong Yu},
      year={2018},
      eprint={1802.01886},
      archivePrefix={arXiv},
      primaryClass={cs.CL},
      url={https://arxiv.org/abs/1802.01886}, 
}

@misc{poole2019variationalboundsmutualinformation,
      title={On Variational Bounds of Mutual Information}, 
      author={Ben Poole and Sherjil Ozair and Aaron van den Oord and Alexander A. Alemi and George Tucker},
      year={2019},
      eprint={1905.06922},
      archivePrefix={arXiv},
      primaryClass={cs.LG},
      url={https://arxiv.org/abs/1905.06922}, 
}

@misc{lambert2025tulu3pushingfrontiers,
      title={Tulu 3: Pushing Frontiers in Open Language Model Post-Training}, 
      author={Nathan Lambert and Jacob Morrison and Valentina Pyatkin and Shengyi Huang and Hamish Ivison and Faeze Brahman and Lester James V. Miranda and Alisa Liu and Nouha Dziri and Shane Lyu and Yuling Gu and Saumya Malik and Victoria Graf and Jena D. Hwang and Jiangjiang Yang and Ronan Le Bras and Oyvind Tafjord and Chris Wilhelm and Luca Soldaini and Noah A. Smith and Yizhong Wang and Pradeep Dasigi and Hannaneh Hajishirzi},
      year={2025},
      eprint={2411.15124},
      archivePrefix={arXiv},
      primaryClass={cs.CL},
      url={https://arxiv.org/abs/2411.15124}, 
}

@misc{shao2024deepseekmathpushinglimitsmathematical,
      title={DeepSeekMath: Pushing the Limits of Mathematical Reasoning in Open Language Models}, 
      author={Zhihong Shao and Peiyi Wang and Qihao Zhu and Runxin Xu and Junxiao Song and Xiao Bi and Haowei Zhang and Mingchuan Zhang and Y. K. Li and Y. Wu and Daya Guo},
      year={2024},
      eprint={2402.03300},
      archivePrefix={arXiv},
      primaryClass={cs.CL},
      url={https://arxiv.org/abs/2402.03300}, 
}

@misc{sorensen2024roadmappluralisticalignment,
      title={A Roadmap to Pluralistic Alignment}, 
      author={Taylor Sorensen and Jared Moore and Jillian Fisher and Mitchell Gordon and Niloofar Mireshghallah and Christopher Michael Rytting and Andre Ye and Liwei Jiang and Ximing Lu and Nouha Dziri and Tim Althoff and Yejin Choi},
      year={2024},
      eprint={2402.05070},
      archivePrefix={arXiv},
      primaryClass={cs.AI},
      url={https://arxiv.org/abs/2402.05070}, 
}

@misc{llama3.2,
title={Llama 3.2: Revolutionizing edge AI and vision with open, customizable models},
author={MetaAI},
url={https://ai.meta.com/blog/llama-3-2-connect-2024-vision-edge-mobile-devices/},
year={2024},
}

@misc{qwen2025qwen25technicalreport,
      title={Qwen2.5 Technical Report}, 
      author={Qwen and : and An Yang and Baosong Yang and Beichen Zhang and Binyuan Hui and Bo Zheng and Bowen Yu and Chengyuan Li and Dayiheng Liu and Fei Huang and Haoran Wei and Huan Lin and Jian Yang and Jianhong Tu and Jianwei Zhang and Jianxin Yang and Jiaxi Yang and Jingren Zhou and Junyang Lin and Kai Dang and Keming Lu and Keqin Bao and Kexin Yang and Le Yu and Mei Li and Mingfeng Xue and Pei Zhang and Qin Zhu and Rui Men and Runji Lin and Tianhao Li and Tianyi Tang and Tingyu Xia and Xingzhang Ren and Xuancheng Ren and Yang Fan and Yang Su and Yichang Zhang and Yu Wan and Yuqiong Liu and Zeyu Cui and Zhenru Zhang and Zihan Qiu},
      year={2025},
      eprint={2412.15115},
      archivePrefix={arXiv},
      primaryClass={cs.CL},
      url={https://arxiv.org/abs/2412.15115}, 
}

@misc{hu2024openrlhfeasytousescalablehighperformance,
      title={OpenRLHF: An Easy-to-use, Scalable and High-performance RLHF Framework}, 
      author={Jian Hu and Xibin Wu and Wei Shen and Jason Klein Liu and Zilin Zhu and Weixun Wang and Songlin Jiang and Haoran Wang and Hao Chen and Bin Chen and Weikai Fang and Xianyu and Yu Cao and Haotian Xu and Yiming Liu},
      year={2025},
      eprint={2405.11143},
      archivePrefix={arXiv},
      primaryClass={cs.AI},
      url={https://arxiv.org/abs/2405.11143}, 
}

@misc{schulman2017proximalpolicyoptimizationalgorithms,
      title={Proximal Policy Optimization Algorithms}, 
      author={John Schulman and Filip Wolski and Prafulla Dhariwal and Alec Radford and Oleg Klimov},
      year={2017},
      eprint={1707.06347},
      archivePrefix={arXiv},
      primaryClass={cs.LG},
      url={https://arxiv.org/abs/1707.06347}, 
}

@misc{huang2025mathperturbbenchmarkingllmsmath,
      title={MATH-Perturb: Benchmarking LLMs' Math Reasoning Abilities against Hard Perturbations}, 
      author={Kaixuan Huang and Jiacheng Guo and Zihao Li and Xiang Ji and Jiawei Ge and Wenzhe Li and Yingqing Guo and Tianle Cai and Hui Yuan and Runzhe Wang and Yue Wu and Ming Yin and Shange Tang and Yangsibo Huang and Chi Jin and Xinyun Chen and Chiyuan Zhang and Mengdi Wang},
      year={2025},
      eprint={2502.06453},
      archivePrefix={arXiv},
      primaryClass={cs.LG},
      url={https://arxiv.org/abs/2502.06453}, 
}

@misc{selfinstruct,
      title={Self-Instruct: Aligning Language Models with Self-Generated Instructions}, 
      author={Yizhong Wang and Yeganeh Kordi and Swaroop Mishra and Alisa Liu and Noah A. Smith and Daniel Khashabi and Hannaneh Hajishirzi},
      year={2023},
      eprint={2212.10560},
      archivePrefix={arXiv},
      primaryClass={cs.CL},
      url={https://arxiv.org/abs/2212.10560}, 
}

@misc{mtbench,
      title={Judging LLM-as-a-Judge with MT-Bench and Chatbot Arena}, 
      author={Lianmin Zheng and Wei-Lin Chiang and Ying Sheng and Siyuan Zhuang and Zhanghao Wu and Yonghao Zhuang and Zi Lin and Zhuohan Li and Dacheng Li and Eric P. Xing and Hao Zhang and Joseph E. Gonzalez and Ion Stoica},
      year={2023},
      eprint={2306.05685},
      archivePrefix={arXiv},
      primaryClass={cs.CL},
      url={https://arxiv.org/abs/2306.05685}, 
}

@misc{koala,
title={Koala Evaluation Set},
author={Arnav Gudibande},
url={https://github.com/arnav-gudibande/koala-test-set},
year={2023}
}

@misc{flask,
      title={FLASK: Fine-grained Language Model Evaluation based on Alignment Skill Sets}, 
      author={Seonghyeon Ye and Doyoung Kim and Sungdong Kim and Hyeonbin Hwang and Seungone Kim and Yongrae Jo and James Thorne and Juho Kim and Minjoon Seo},
      year={2024},
      eprint={2307.10928},
      archivePrefix={arXiv},
      primaryClass={cs.CL},
      url={https://arxiv.org/abs/2307.10928}, 
}

@misc{poddar2024personalizingreinforcementlearninghuman,
      title={Personalizing Reinforcement Learning from Human Feedback with Variational Preference Learning}, 
      author={Sriyash Poddar and Yanming Wan and Hamish Ivison and Abhishek Gupta and Natasha Jaques},
      year={2024},
      eprint={2408.10075},
      archivePrefix={arXiv},
      primaryClass={cs.LG},
      url={https://arxiv.org/abs/2408.10075}, 
}

@misc{chen2024selfplayfinetuningconvertsweak,
      title={Self-Play Fine-Tuning Converts Weak Language Models to Strong Language Models}, 
      author={Zixiang Chen and Yihe Deng and Huizhuo Yuan and Kaixuan Ji and Quanquan Gu},
      year={2024},
      eprint={2401.01335},
      archivePrefix={arXiv},
      primaryClass={cs.LG},
      url={https://arxiv.org/abs/2401.01335}, 
}

@misc{stiennon2022learningsummarizehumanfeedback,
      title={Learning to summarize from human feedback}, 
      author={Nisan Stiennon and Long Ouyang and Jeff Wu and Daniel M. Ziegler and Ryan Lowe and Chelsea Voss and Alec Radford and Dario Amodei and Paul Christiano},
      year={2022},
      eprint={2009.01325},
      archivePrefix={arXiv},
      primaryClass={cs.CL},
      url={https://arxiv.org/abs/2009.01325}, 
}

@misc{dong2025stpselfplayllmtheorem,
      title={STP: Self-play LLM Theorem Provers with Iterative Conjecturing and Proving}, 
      author={Kefan Dong and Tengyu Ma},
      year={2025},
      eprint={2502.00212},
      archivePrefix={arXiv},
      primaryClass={cs.LG},
      url={https://arxiv.org/abs/2502.00212}, 
}

@misc{doosterlinck2024anchoredpreferenceoptimizationcontrastive,
      title={Anchored Preference Optimization and Contrastive Revisions: Addressing Underspecification in Alignment}, 
      author={Karel D'Oosterlinck and Winnie Xu and Chris Develder and Thomas Demeester and Amanpreet Singh and Christopher Potts and Douwe Kiela and Shikib Mehri},
      year={2024},
      eprint={2408.06266},
      archivePrefix={arXiv},
      primaryClass={cs.LG},
      url={https://arxiv.org/abs/2408.06266}, 
}

@misc{xiao2025connectionimitationlearningrlhf,
      title={On a Connection Between Imitation Learning and RLHF}, 
      author={Teng Xiao and Yige Yuan and Mingxiao Li and Zhengyu Chen and Vasant G Honavar},
      year={2025},
      eprint={2503.05079},
      archivePrefix={arXiv},
      primaryClass={cs.LG},
      url={https://arxiv.org/abs/2503.05079}, 
}

@misc{yang2024weaktostrongreasoning,
      title={Weak-to-Strong Reasoning}, 
      author={Yuqing Yang and Yan Ma and Pengfei Liu},
      year={2024},
      eprint={2407.13647},
      archivePrefix={arXiv},
      primaryClass={cs.CL},
      url={https://arxiv.org/abs/2407.13647}, 
}

@misc{zhu2025weaktostrongpreferenceoptimizationstealing,
      title={Weak-to-Strong Preference Optimization: Stealing Reward from Weak Aligned Model}, 
      author={Wenhong Zhu and Zhiwei He and Xiaofeng Wang and Pengfei Liu and Rui Wang},
      year={2025},
      eprint={2410.18640},
      archivePrefix={arXiv},
      primaryClass={cs.CL},
      url={https://arxiv.org/abs/2410.18640}, 
}

@misc{zhang2025interplaypretrainingmidtrainingrl,
      title={On the Interplay of Pre-Training, Mid-Training, and RL on Reasoning Language Models}, 
      author={Charlie Zhang and Graham Neubig and Xiang Yue},
      year={2025},
      eprint={2512.07783},
      archivePrefix={arXiv},
      primaryClass={cs.CL},
      url={https://arxiv.org/abs/2512.07783}, 
}

@misc{oord2019representationlearningcontrastivepredictive,
      title={Representation Learning with Contrastive Predictive Coding}, 
      author={Aaron van den Oord and Yazhe Li and Oriol Vinyals},
      year={2019},
      eprint={1807.03748},
      archivePrefix={arXiv},
      primaryClass={cs.LG},
      url={https://arxiv.org/abs/1807.03748}, 
}

@misc{clark2018thinksolvedquestionanswering,
      title={Think you have Solved Question Answering? Try ARC, the AI2 Reasoning Challenge}, 
      author={Peter Clark and Isaac Cowhey and Oren Etzioni and Tushar Khot and Ashish Sabharwal and Carissa Schoenick and Oyvind Tafjord},
      year={2018},
      eprint={1803.05457},
      archivePrefix={arXiv},
      primaryClass={cs.AI},
      url={https://arxiv.org/abs/1803.05457}, 
}

@misc{gu2025olmesstandardlanguagemodel,
      title={OLMES: A Standard for Language Model Evaluations}, 
      author={Yuling Gu and Oyvind Tafjord and Bailey Kuehl and Dany Haddad and Jesse Dodge and Hannaneh Hajishirzi},
      year={2025},
      eprint={2406.08446},
      archivePrefix={arXiv},
      primaryClass={cs.CL},
      url={https://arxiv.org/abs/2406.08446}, 
}

@misc{patel2021nlpmodelsreallyable,
      title={Are NLP Models really able to Solve Simple Math Word Problems?}, 
      author={Arkil Patel and Satwik Bhattamishra and Navin Goyal},
      year={2021},
      eprint={2103.07191},
      archivePrefix={arXiv},
      primaryClass={cs.CL},
      url={https://arxiv.org/abs/2103.07191}, 
}

@misc{cobbe2021trainingverifierssolvemath,
      title={Training Verifiers to Solve Math Word Problems}, 
      author={Karl Cobbe and Vineet Kosaraju and Mohammad Bavarian and Mark Chen and Heewoo Jun and Lukasz Kaiser and Matthias Plappert and Jerry Tworek and Jacob Hilton and Reiichiro Nakano and Christopher Hesse and John Schulman},
      year={2021},
      eprint={2110.14168},
      archivePrefix={arXiv},
      primaryClass={cs.LG},
      url={https://arxiv.org/abs/2110.14168}, 
}

@misc{zhang2025cultivatingpluralismalgorithmicmonoculture,
      title={Cultivating Pluralism In Algorithmic Monoculture: The Community Alignment Dataset}, 
      author={Lily Hong Zhang and Smitha Milli and Karen Jusko and Jonathan Smith and Brandon Amos and Wassim Bouaziz and Manon Revel and Jack Kussman and Yasha Sheynin and Lisa Titus and Bhaktipriya Radharapu and Jane Yu and Vidya Sarma and Kris Rose and Maximilian Nickel},
      year={2025},
      eprint={2507.09650},
      archivePrefix={arXiv},
      primaryClass={cs.LG},
      url={https://arxiv.org/abs/2507.09650}, 
}

@article{krizhevsky2012imagenet,
  title={Imagenet classification with deep convolutional neural networks},
  author={Krizhevsky, Alex and Sutskever, Ilya and Hinton, Geoffrey E},
  journal={Advances in neural information processing systems},
  volume={25},
  year={2012}
}

@misc{geng2025deltalearninghypothesispreference,
      title={The Delta Learning Hypothesis: Preference Tuning on Weak Data can Yield Strong Gains}, 
      author={Scott Geng and Hamish Ivison and Chun-Liang Li and Maarten Sap and Jerry Li and Ranjay Krishna and Pang Wei Koh},
      year={2025},
      eprint={2507.06187},
      archivePrefix={arXiv},
      primaryClass={cs.AI},
      url={https://arxiv.org/abs/2507.06187}, 
}

@misc{yao2024varyingshadeswrongaligning,
      title={Varying Shades of Wrong: Aligning LLMs with Wrong Answers Only}, 
      author={Jihan Yao and Wenxuan Ding and Shangbin Feng and Lucy Lu Wang and Yulia Tsvetkov},
      year={2024},
      eprint={2410.11055},
      archivePrefix={arXiv},
      primaryClass={cs.CL},
      url={https://arxiv.org/abs/2410.11055}, 
}

@misc{lv2025hiddenlinkrlhfcontrastive,
      title={The Hidden Link Between RLHF and Contrastive Learning}, 
      author={Xufei Lv and Kehai Chen and Haoyuan Sun and Xuefeng Bai and Min Zhang and Houde Liu and Kehai Chen},
      year={2025},
      eprint={2506.22578},
      archivePrefix={arXiv},
      primaryClass={cs.LG},
      url={https://arxiv.org/abs/2506.22578}, 
}

@misc{nam2026learningsummarizeuserinformation,
      title={Learning to summarize user information for personalized reinforcement learning from human feedback}, 
      author={Hyunji Nam and Yanming Wan and Mickel Liu and Peter Ahnn and Jianxun Lian and Natasha Jaques},
      year={2026},
      eprint={2507.13579},
      archivePrefix={arXiv},
      primaryClass={cs.LG},
      url={https://arxiv.org/abs/2507.13579}, 
}

@misc{abdulhai2026llmsdistortwrittenlanguage,
      title={How LLMs Distort Our Written Language}, 
      author={Marwa Abdulhai and Isadora White and Yanming Wan and Ibrahim Qureshi and Joel Leibo and Max Kleiman-Weiner and Natasha Jaques},
      year={2026},
      eprint={2603.18161},
      archivePrefix={arXiv},
      primaryClass={cs.CL},
      url={https://arxiv.org/abs/2603.18161}, 
}

@misc{zhao2025llmsrecognizepreferencesevaluating,
      title={Do LLMs Recognize Your Preferences? Evaluating Personalized Preference Following in LLMs}, 
      author={Siyan Zhao and Mingyi Hong and Yang Liu and Devamanyu Hazarika and Kaixiang Lin},
      year={2025},
      eprint={2502.09597},
      archivePrefix={arXiv},
      primaryClass={cs.LG},
      url={https://arxiv.org/abs/2502.09597}, 
}

@misc{li2025personalizedreasoningjustintimepersonalization,
      title={PrefDisco: Benchmarking Proactive Personalized Reasoning}, 
      author={Shuyue Stella Li and Avinandan Bose and Faeze Brahman and Simon Shaolei Du and Pang Wei Koh and Maryam Fazel and Yulia Tsvetkov},
      year={2026},
      eprint={2510.00177},
      archivePrefix={arXiv},
      primaryClass={cs.CL},
      url={https://arxiv.org/abs/2510.00177}, 
}

@misc{jiang2025knowmerespondme,
      title={Know Me, Respond to Me: Benchmarking LLMs for Dynamic User Profiling and Personalized Responses at Scale}, 
      author={Bowen Jiang and Zhuoqun Hao and Young-Min Cho and Bryan Li and Yuan Yuan and Sihao Chen and Lyle Ungar and Camillo J. Taylor and Dan Roth},
      year={2025},
      eprint={2504.14225},
      archivePrefix={arXiv},
      primaryClass={cs.CL},
      url={https://arxiv.org/abs/2504.14225}, 
}

@misc{liu2025sharedlowrankadaptationapproach,
      title={A Shared Low-Rank Adaptation Approach to Personalized RLHF}, 
      author={Renpu Liu and Peng Wang and Donghao Li and Cong Shen and Jing Yang},
      year={2025},
      eprint={2503.19201},
      archivePrefix={arXiv},
      primaryClass={cs.LG},
      url={https://arxiv.org/abs/2503.19201}, 
}

@misc{kim2025cupidevaluatingpersonalizedcontextualized,
      title={CUPID: Evaluating Personalized and Contextualized Alignment of LLMs from Interactions}, 
      author={Tae Soo Kim and Yoonjoo Lee and Yoonah Park and Jiho Kim and Young-Ho Kim and Juho Kim},
      year={2025},
      eprint={2508.01674},
      archivePrefix={arXiv},
      primaryClass={cs.CL},
      url={https://arxiv.org/abs/2508.01674}, 
}

@misc{rafailov2024directpreferenceoptimizationlanguage,
      title={Direct Preference Optimization: Your Language Model is Secretly a Reward Model}, 
      author={Rafael Rafailov and Archit Sharma and Eric Mitchell and Stefano Ermon and Christopher D. Manning and Chelsea Finn},
      year={2024},
      eprint={2305.18290},
      archivePrefix={arXiv},
      primaryClass={cs.LG},
      url={https://arxiv.org/abs/2305.18290}, 
}

@misc{ouyang2022traininglanguagemodelsfollow,
      title={Training language models to follow instructions with human feedback}, 
      author={Long Ouyang and Jeff Wu and Xu Jiang and Diogo Almeida and Carroll L. Wainwright and Pamela Mishkin and Chong Zhang and Sandhini Agarwal and Katarina Slama and Alex Ray and John Schulman and Jacob Hilton and Fraser Kelton and Luke Miller and Maddie Simens and Amanda Askell and Peter Welinder and Paul Christiano and Jan Leike and Ryan Lowe},
      year={2022},
      eprint={2203.02155},
      archivePrefix={arXiv},
      primaryClass={cs.CL},
      url={https://arxiv.org/abs/2203.02155}, 
}

@misc{dubois2025lengthcontrolledalpacaevalsimpleway,
      title={Length-Controlled AlpacaEval: A Simple Way to Debias Automatic Evaluators}, 
      author={Yann Dubois and Balázs Galambosi and Percy Liang and Tatsunori B. Hashimoto},
      year={2025},
      eprint={2404.04475},
      archivePrefix={arXiv},
      primaryClass={cs.LG},
      url={https://arxiv.org/abs/2404.04475}, 
}

@misc{hu2021loralowrankadaptationlarge,
      title={LoRA: Low-Rank Adaptation of Large Language Models}, 
      author={Edward J. Hu and Yelong Shen and Phillip Wallis and Zeyuan Allen-Zhu and Yuanzhi Li and Shean Wang and Lu Wang and Weizhu Chen},
      year={2021},
      eprint={2106.09685},
      archivePrefix={arXiv},
      primaryClass={cs.CL},
      url={https://arxiv.org/abs/2106.09685}, 
}

@misc{touvron2023llamaopenefficientfoundation,
      title={LLaMA: Open and Efficient Foundation Language Models}, 
      author={Hugo Touvron and Thibaut Lavril and Gautier Izacard and Xavier Martinet and Marie-Anne Lachaux and Timothée Lacroix and Baptiste Rozière and Naman Goyal and Eric Hambro and Faisal Azhar and Aurelien Rodriguez and Armand Joulin and Edouard Grave and Guillaume Lample},
      year={2023},
      eprint={2302.13971},
      archivePrefix={arXiv},
      primaryClass={cs.CL},
      url={https://arxiv.org/abs/2302.13971}, 
}

@misc{huang2024largelanguagemodelsselfcorrect,
      title={Large Language Models Cannot Self-Correct Reasoning Yet}, 
      author={Jie Huang and Xinyun Chen and Swaroop Mishra and Huaixiu Steven Zheng and Adams Wei Yu and Xinying Song and Denny Zhou},
      year={2024},
      eprint={2310.01798},
      archivePrefix={arXiv},
      primaryClass={cs.CL},
      url={https://arxiv.org/abs/2310.01798}, 
}

@misc{das2025activepreferenceoptimizationsample,
      title={Active Preference Optimization for Sample Efficient RLHF}, 
      author={Nirjhar Das and Souradip Chakraborty and Aldo Pacchiano and Sayak Ray Chowdhury},
      year={2025},
      eprint={2402.10500},
      archivePrefix={arXiv},
      primaryClass={cs.LG},
      url={https://arxiv.org/abs/2402.10500}, 
}

@misc{mazoure2022contrastivevaluelearningimplicit,
      title={Contrastive Value Learning: Implicit Models for Simple Offline RL}, 
      author={Bogdan Mazoure and Benjamin Eysenbach and Ofir Nachum and Jonathan Tompson},
      year={2022},
      eprint={2211.02100},
      archivePrefix={arXiv},
      primaryClass={cs.LG},
      url={https://arxiv.org/abs/2211.02100}, 
}

@misc{samokhin2025randomdirectpreferenceoptimization,
      title={Random Direct Preference Optimization for Radiography Report Generation}, 
      author={Valentin Samokhin and Boris Shirokikh and Mikhail Goncharov and Dmitriy Umerenkov and Maksim Bobrin and Ivan Oseledets and Dmitry Dylov and Mikhail Belyaev},
      year={2025},
      eprint={2509.21351},
      archivePrefix={arXiv},
      primaryClass={cs.CV},
      url={https://arxiv.org/abs/2509.21351}, 
}

@misc{yin2024relativepreferenceoptimizationenhancing,
      title={Relative Preference Optimization: Enhancing LLM Alignment through Contrasting Responses across Identical and Diverse Prompts}, 
      author={Yueqin Yin and Zhendong Wang and Yi Gu and Hai Huang and Weizhu Chen and Mingyuan Zhou},
      year={2024},
      eprint={2402.10958},
      archivePrefix={arXiv},
      primaryClass={cs.CL},
      url={https://arxiv.org/abs/2402.10958}, 
}

@misc{xu2024automaticpairconstructioncontrastive,
      title={Automatic Pair Construction for Contrastive Post-training}, 
      author={Canwen Xu and Corby Rosset and Ethan C. Chau and Luciano Del Corro and Shweti Mahajan and Julian McAuley and Jennifer Neville and Ahmed Hassan Awadallah and Nikhil Rao},
      year={2024},
      eprint={2310.02263},
      archivePrefix={arXiv},
      primaryClass={cs.CL},
      url={https://arxiv.org/abs/2310.02263}, 
}

@misc{hejna2024contrastivepreferencelearninglearning,
      title={Contrastive Preference Learning: Learning from Human Feedback without RL}, 
      author={Joey Hejna and Rafael Rafailov and Harshit Sikchi and Chelsea Finn and Scott Niekum and W. Bradley Knox and Dorsa Sadigh},
      year={2024},
      eprint={2310.13639},
      archivePrefix={arXiv},
      primaryClass={cs.LG},
      url={https://arxiv.org/abs/2310.13639}, 
}

@misc{eysenbach2023contrastivelearninggoalconditionedreinforcement,
      title={Contrastive Learning as Goal-Conditioned Reinforcement Learning}, 
      author={Benjamin Eysenbach and Tianjun Zhang and Ruslan Salakhutdinov and Sergey Levine},
      year={2023},
      eprint={2206.07568},
      archivePrefix={arXiv},
      primaryClass={cs.LG},
      url={https://arxiv.org/abs/2206.07568}, 
}

@misc{chen2020simpleframeworkcontrastivelearning,
      title={A Simple Framework for Contrastive Learning of Visual Representations}, 
      author={Ting Chen and Simon Kornblith and Mohammad Norouzi and Geoffrey Hinton},
      year={2020},
      eprint={2002.05709},
      archivePrefix={arXiv},
      primaryClass={cs.LG},
      url={https://arxiv.org/abs/2002.05709}, 
}

@misc{qi2025difficultybasedpreferencedataselection,
      title={Difficulty-Based Preference Data Selection by DPO Implicit Reward Gap}, 
      author={Xuan Qi and Rongwu Xu and Zhijing Jin},
      year={2025},
      eprint={2508.04149},
      archivePrefix={arXiv},
      primaryClass={cs.CL},
      url={https://arxiv.org/abs/2508.04149}, 
}

@misc{dwaracherla2024efficientexplorationllms,
      title={Efficient Exploration for LLMs}, 
      author={Vikranth Dwaracherla and Seyed Mohammad Asghari and Botao Hao and Benjamin Van Roy},
      year={2024},
      eprint={2402.00396},
      archivePrefix={arXiv},
      primaryClass={cs.LG},
      url={https://arxiv.org/abs/2402.00396}, 
}

@misc{gou2025mixedpreferenceoptimizationreinforcement,
      title={Mixed Preference Optimization: Reinforcement Learning with Data Selection and Better Reference Model}, 
      author={Qi Gou and Cam-Tu Nguyen},
      year={2025},
      eprint={2403.19443},
      archivePrefix={arXiv},
      primaryClass={cs.CL},
      url={https://arxiv.org/abs/2403.19443}, 
}

@misc{anderson2018visionandlanguagenavigationinterpretingvisuallygrounded,
      title={Vision-and-Language Navigation: Interpreting visually-grounded navigation instructions in real environments}, 
      author={Peter Anderson and Qi Wu and Damien Teney and Jake Bruce and Mark Johnson and Niko Sünderhauf and Ian Reid and Stephen Gould and Anton van den Hengel},
      year={2018},
      eprint={1711.07280},
      archivePrefix={arXiv},
      primaryClass={cs.CV},
      url={https://arxiv.org/abs/1711.07280}, 
}

@misc{tyen2024llmsreasoningerrorscorrect,
      title={LLMs cannot find reasoning errors, but can correct them given the error location}, 
      author={Gladys Tyen and Hassan Mansoor and Victor Cărbune and Peter Chen and Tony Mak},
      year={2024},
      eprint={2311.08516},
      archivePrefix={arXiv},
      primaryClass={cs.AI},
      url={https://arxiv.org/abs/2311.08516}, 
}

@misc{lee2024rlaifvsrlhfscaling,
      title={RLAIF vs. RLHF: Scaling Reinforcement Learning from Human Feedback with AI Feedback}, 
      author={Harrison Lee and Samrat Phatale and Hassan Mansoor and Thomas Mesnard and Johan Ferret and Kellie Lu and Colton Bishop and Ethan Hall and Victor Carbune and Abhinav Rastogi and Sushant Prakash},
      year={2024},
      eprint={2309.00267},
      archivePrefix={arXiv},
      primaryClass={cs.CL},
      url={https://arxiv.org/abs/2309.00267}, 
}

@misc{hosseini2024vstartrainingverifiersselftaught,
      title={V-STaR: Training Verifiers for Self-Taught Reasoners}, 
      author={Arian Hosseini and Xingdi Yuan and Nikolay Malkin and Aaron Courville and Alessandro Sordoni and Rishabh Agarwal},
      year={2024},
      eprint={2402.06457},
      archivePrefix={arXiv},
      primaryClass={cs.LG},
      url={https://arxiv.org/abs/2402.06457}, 
}

@misc{yuan2023scalingrelationshiplearningmathematical,
      title={Scaling Relationship on Learning Mathematical Reasoning with Large Language Models}, 
      author={Zheng Yuan and Hongyi Yuan and Chengpeng Li and Guanting Dong and Keming Lu and Chuanqi Tan and Chang Zhou and Jingren Zhou},
      year={2023},
      eprint={2308.01825},
      archivePrefix={arXiv},
      primaryClass={cs.CL},
      url={https://arxiv.org/abs/2308.01825}, 
}

@misc{zelikman2022starbootstrappingreasoningreasoning,
      title={STaR: Bootstrapping Reasoning With Reasoning}, 
      author={Eric Zelikman and Yuhuai Wu and Jesse Mu and Noah D. Goodman},
      year={2022},
      eprint={2203.14465},
      archivePrefix={arXiv},
      primaryClass={cs.LG},
      url={https://arxiv.org/abs/2203.14465}, 
}

@misc{singh2024humandatascalingselftraining,
      title={Beyond Human Data: Scaling Self-Training for Problem-Solving with Language Models}, 
      author={Avi Singh and John D. Co-Reyes and Rishabh Agarwal and Ankesh Anand and Piyush Patil and Xavier Garcia and Peter J. Liu and James Harrison and Jaehoon Lee and Kelvin Xu and Aaron Parisi and Abhishek Kumar and Alex Alemi and Alex Rizkowsky and Azade Nova and Ben Adlam and Bernd Bohnet and Gamaleldin Elsayed and Hanie Sedghi and Igor Mordatch and Isabelle Simpson and Izzeddin Gur and Jasper Snoek and Jeffrey Pennington and Jiri Hron and Kathleen Kenealy and Kevin Swersky and Kshiteej Mahajan and Laura Culp and Lechao Xiao and Maxwell L. Bileschi and Noah Constant and Roman Novak and Rosanne Liu and Tris Warkentin and Yundi Qian and Yamini Bansal and Ethan Dyer and Behnam Neyshabur and Jascha Sohl-Dickstein and Noah Fiedel},
      year={2024},
      eprint={2312.06585},
      archivePrefix={arXiv},
      primaryClass={cs.LG},
      url={https://arxiv.org/abs/2312.06585}, 
}

@misc{ulmer2024bootstrappingllmbasedtaskorienteddialogue,
      title={Bootstrapping LLM-based Task-Oriented Dialogue Agents via Self-Talk}, 
      author={Dennis Ulmer and Elman Mansimov and Kaixiang Lin and Justin Sun and Xibin Gao and Yi Zhang},
      year={2024},
      eprint={2401.05033},
      archivePrefix={arXiv},
      primaryClass={cs.CL},
      url={https://arxiv.org/abs/2401.05033}, 
}

@misc{shridhar2023artllmrefinementask,
      title={The ART of LLM Refinement: Ask, Refine, and Trust}, 
      author={Kumar Shridhar and Koustuv Sinha and Andrew Cohen and Tianlu Wang and Ping Yu and Ram Pasunuru and Mrinmaya Sachan and Jason Weston and Asli Celikyilmaz},
      year={2023},
      eprint={2311.07961},
      archivePrefix={arXiv},
      primaryClass={cs.CL},
      url={https://arxiv.org/abs/2311.07961}, 
}

@misc{madaan2023selfrefineiterativerefinementselffeedback,
      title={Self-Refine: Iterative Refinement with Self-Feedback}, 
      author={Aman Madaan and Niket Tandon and Prakhar Gupta and Skyler Hallinan and Luyu Gao and Sarah Wiegreffe and Uri Alon and Nouha Dziri and Shrimai Prabhumoye and Yiming Yang and Shashank Gupta and Bodhisattwa Prasad Majumder and Katherine Hermann and Sean Welleck and Amir Yazdanbakhsh and Peter Clark},
      year={2023},
      eprint={2303.17651},
      archivePrefix={arXiv},
      primaryClass={cs.CL},
      url={https://arxiv.org/abs/2303.17651}, 
}

@misc{chen2024iteraligniterativeconstitutionalalignment,
      title={IterAlign: Iterative Constitutional Alignment of Large Language Models}, 
      author={Xiusi Chen and Hongzhi Wen and Sreyashi Nag and Chen Luo and Qingyu Yin and Ruirui Li and Zheng Li and Wei Wang},
      year={2024},
      eprint={2403.18341},
      archivePrefix={arXiv},
      primaryClass={cs.CL},
      url={https://arxiv.org/abs/2403.18341}, 
}

@misc{liang2024isheepselfalignmentllmscratch,
      title={I-SHEEP: Self-Alignment of LLM from Scratch through an Iterative Self-Enhancement Paradigm}, 
      author={Yiming Liang and Ge Zhang and Xingwei Qu and Tianyu Zheng and Jiawei Guo and Xinrun Du and Zhenzhu Yang and Jiaheng Liu and Chenghua Lin and Lei Ma and Wenhao Huang and Jiajun Zhang},
      year={2024},
      eprint={2408.08072},
      archivePrefix={arXiv},
      primaryClass={cs.CL},
      url={https://arxiv.org/abs/2408.08072}, 
}
\clearpage
\onecolumn
\appendix
\section{Appendix: Related Work}\label{appendix:related_work}
\textbf{Reinforcement Learning (RL) fine-tuning of LLMs.}
Post-training with RL has proven effective for tasks that target a model's \emph{edge of competence}, i.e., tasks that are difficult but within the pre-training data distribution~\citep{zhang2025interplaypretrainingmidtrainingrl}. Two popular frameworks are: RL with Verifiable Rewards (RLVR)~\citep{lambert2025tulu3pushingfrontiers, Guo_2025} and RL with Human Feedback (RLHF)~\citep{jaques2019way,stiennon2022learningsummarizehumanfeedback, ouyang2022traininglanguagemodelsfollow} based on whether the reward is verifiable or estimated from human preference data. Optimization can be implemented with different RL algorithms, including Proximal Policy Optimization (PPO)~\citep{schulman2017proximalpolicyoptimizationalgorithms}, Group Relative Policy Optimization (GRPO)~\citep{shao2024deepseekmathpushinglimitsmathematical}, and Direct Preference Optimization (DPO)~\citep{rafailov2024directpreferenceoptimizationlanguage}.
% has been extensively adopted~\citep{} and studied~\citep{lv2025hiddenlinkrlhfcontrastive, xiao2025connectionimitationlearningrlhf}.

% While iterative refinement and self-feedback has shown success, for example by prompting for feedback and revision~\citep{madaan2023selfrefineiterativerefinementselffeedback, shridhar2023artllmrefinementask} or using in-context learning to progressively build reasoning~\citep{zelikman2022starbootstrappingreasoningreasoning}, ...
\textbf{Self-training and improvement.} RL from AI feedback (RLAIF) reduces the burden of human preference data collection by using an LLM-as-a-judge~\citep{bai2022constitutionalaiharmlessnessai, lee2024rlaifvsrlhfscaling} or self-rewarding models~\citep{yuan2025selfrewardinglanguagemodels}. Other methods use multi-agent reinforcement learning (MARL) to train both the generator and the evaluator, allowing the generator to learn from the evaluator's reward signal~\citep{dong2025stpselfplayllmtheorem, chen2024selfplayfinetuningconvertsweak}. However, these approaches still rely on verifiable rewards or human-annotated data in order to train a reliable evaluator. Similarly, self-supervised training methods often require external verifiers to ensure correctness of the training data or use human-generated rubrics for quality assessment~\citep{singh2024humandatascalingselftraining, ulmer2024bootstrappingllmbasedtaskorienteddialogue, liang2024isheepselfalignmentllmscratch, hosseini2024vstartrainingverifiersselftaught, yuan2023scalingrelationshiplearningmathematical}. While iterative refinement and self-feedback has shown success~\citep{madaan2023selfrefineiterativerefinementselffeedback, shridhar2023artllmrefinementask, zelikman2022starbootstrappingreasoningreasoning}, \citet{huang2024largelanguagemodelsselfcorrect} observe that self-correction can degrade model performance when external feedback or verifiers are not available. ~\citet{tyen2024llmsreasoningerrorscorrect} speculate that this may be due to the model's limited capability to identify mistakes rather than to correct them. We also observe in our experiments that prompting models to self-revise and training on these revisions can cause performance degradation in some tasks.

\textbf{Direct Preference Optimization (DPO) with data augmentation.} Prior works have explored different data generation or selection methods to improve RLHF outcomes~\citep{dwaracherla2024efficientexplorationllms, das2025activepreferenceoptimizationsample, gou2025mixedpreferenceoptimizationreinforcement, qi2025difficultybasedpreferencedataselection, yang2024weaktostrongreasoning, zhu2025weaktostrongpreferenceoptimizationstealing, yao2024varyingshadeswrongaligning, geng2025deltalearninghypothesispreference}. Specifically for DPO, we highlight three data generation strategies from prior work: (1) Pairing suboptimal data for contrastive signals. ~\citet{yao2024varyingshadeswrongaligning} empirically show that DPO works with \emph{less-wrong-over-wrong} pairs, even when the chosen responses are also incorrect. Similarly, ~\citet{xu2024automaticpairconstructioncontrastive, geng2025deltalearninghypothesispreference} show that contrasting outputs from a strong model against those from a weaker model leads to model improvement. (2) Improving data through targeted revision. ~\citet{doosterlinck2024anchoredpreferenceoptimizationcontrastive} propose improving rejected responses along specific dimensions to obtain targeted, higher-quality chosen responses. (3) Bootstrapping from original data by applying perturbations in either positive or negative directions. When ground-truth responses are available, ~\citet{samokhin2025randomdirectpreferenceoptimization} suggest pairing correct responses with random examples from the dataset to create negative pairs. ~\citet{anderson2018visionandlanguagenavigationinterpretingvisuallygrounded} create negative samples by adding semantic perturbations to the original text. ~\citet{yin2024relativepreferenceoptimizationenhancing} expand preference pairs based on semantic similarity of prompts to pair the chosen and rejected responses with additional prompts. While these works provide promising results suggesting benefits of textual data augmentation, they depend on larger models~\citep{xu2024automaticpairconstructioncontrastive, geng2025deltalearninghypothesispreference}, human instructions for revision~\citep{doosterlinck2024anchoredpreferenceoptimizationcontrastive}, or ground-truth data or verifiers to ensure the relative quality of the chosen over the rejected~\citep{samokhin2025randomdirectpreferenceoptimization, yin2024relativepreferenceoptimizationenhancing}. In contrast, we show that models can self-improve without any external supervision or verifiers by contrasting responses conditioned on correct versus incorrect prompts.
% We build on these successes to develop a self-training algorithm that requires only prompts and evaluate it across diverse language modeling tasks, from open-ended text generation to math.

\textbf{Personalization of LLMs.} Despite the success of making LLMs more helpful and performant~\citep{bai2022traininghelpfulharmlessassistant, ouyang2022traininglanguagemodelsfollow}, recent work shows that human preferences are diverse and sometimes even conflicting~\citep{sorensen2024roadmappluralisticalignment, kirk2024prismalignmentdatasetparticipatory, zhang2025cultivatingpluralismalgorithmicmonoculture}. Therefore, preference alignment for LLMs should consider pluralistic or personalized alignment to individual preferences and traits. This capability is also known as \emph{in-context adaption}, as models need to steer their outputs based on in-situ information gathered about users during conversations~\citep{sorensen2025spectrumtuningposttrainingdistributional}. To address this growing interest, many personalization benchmarks, using a mix of synthetic and real-user data, %modeling user interactions with LLM chatbots
have been constructed
%as personalization benchmarks
~\citep{lee2024aligningthousandspreferencesmessage, kim2025cupidevaluatingpersonalizedcontextualized, zhao2025llmsrecognizepreferencesevaluating, jiang2025knowmerespondme, li2025personalizedreasoningjustintimepersonalization}. A number of methods have also been developed to improve personalization, sharing the core idea of learning a user-conditioned policy or preference model that can distinguish among heterogeneous preferences~\citep{li2024personalizedlanguagemodelingpersonalized, liu2025sharedlowrankadaptationapproach, poddar2024personalizingreinforcementlearninghuman, nam2026learningsummarizeuserinformation}). Unlike prior approaches that rely on expert demonstrations or human preference datasets, we show that personalization can be achieved without human supervision by leveraging the model’s intrinsic signals derived from contrastive response pairs. Closest to our work, \citet{fränken2024selfsupervisedalignmentmutualinformation} use an InfoNCE-derived loss for constitution following; we adapt their method to personalization as one of our baselines. Rather than introducing a new loss function, MIPO is built directly on DPO, a widely used post-training algorithm. As a result, MIPO has the practical advantage of reducing to a simple, plug-in data augmentation step before standard DPO training. Moreover, MIPO can be applied flexibly to both personalization and general LLM tasks, showing empirical gains in both, unlike the prior approach which is limited to settings with decoupled prompts and constitutions. 

\section{Appendix: Results}

\begin{table}[h]
% \centering
% \begin{minipage}{0.5\textwidth}
    \caption{\textbf{Self-BLEU-4~\cite{zhu2018texygenbenchmarkingplatformtext} from pre- and post-training in (i) CA Community Alignment (CA), (ii) PRISM, (iii) Multi-bench (MB).} We report the mean from 3 seeds for all trained models, except Qwen-7B. Lower values mean higher diversity (\textcolor{darkgreen}{\ding{51}} indicates diversity improvement).} 
\label{tab:self_bleu_diversity}
\centering
\adjustbox{width=0.7\textwidth}{
\begin{tabular}{lccccc}
\toprule
Model & CA & PRISM & MB  & Avg. \\
\midrule
\rowcolor{basebg}
\textbf{Llama-3.2-1B-Instruct} (Personalized-Prompting) & 0.393 & \textbf{0.356} & 0.510 & 0.420 \\
\quad + SFT & \textbf{0.362} & 0.384 & 0.535 & 0.427 \\
\rowcolor{oursbg}
\quad + \textbf{MIPO-Personalized} & 0.363 & 0.365 & \textbf{0.450} & \textbf{0.393 \textcolor{darkgreen}{\ding{51}}}\\
\midrule
\rowcolor{basebg}
\textbf{Llama-3.2-3B-Instruct} (Personalized-Prompting) & \textbf{0.311} & \textbf{0.274} & 0.551 & 0.379 \\
\quad + SFT & 0.330 & 0.297 & 0.539 & 0.389 \\
\rowcolor{oursbg}
\quad + \textbf{MIPO-Personalized} & 0.329 & 0.288 & \textbf{0.495} & \textbf{0.371 \textcolor{darkgreen}{\ding{51}}}\\
\midrule
\textbf{Qwen2.5-1.5B-Instruct} (Personalized-Prompting) & 0.208 & 0.167 & 0.554 & 0.310 \\
\quad + SFT & 0.231 & 0.178 & 0.538 & 0.316 \\
\rowcolor{oursbg}
\quad + \textbf{MIPO-Personalized} & \textbf{0.195} & \textbf{0.150} & \textbf{0.423} & \textbf{0.256 \textcolor{darkgreen}{\ding{51}}}\\
\midrule
\rowcolor{basebg}
\textbf{Qwen2.5-3B-Instruct} (Personalized-Prompting) & 0.200 & 0.156 & 0.581 & 0.312 \\
\quad + SFT & 0.225 & 0.158 & 0.564 & 0.316 \\
\rowcolor{oursbg}
\quad + \textbf{MIPO-Personalized} & \textbf{0.188} & \textbf{0.144} & \textbf{0.483} & \textbf{0.272 \textcolor{darkgreen}{\ding{51}}} \\
\rowcolor{basebg}
\midrule\textbf{Qwen2.5-7B-Instruct} (Personalized-Prompting) & 0.197 & 0.158 & 0.575 & 0.310 \\
\quad + SFT & 0.211 & \textbf{0.160} & 0.583 & 0.318 \\
\rowcolor{oursbg}
\quad + \textbf{MIPO-Personalized} & \textbf{0.193} & 0.163 & \textbf{0.570} & \textbf{0.309 \textcolor{darkgreen}{\ding{51}}}\\
\bottomrule
\end{tabular}}
\vspace{-0.2cm}
\end{table}

\begin{table}[H]
% \vspace{-0.5cm}
% \centering
% \begin{minipage}{0.5\textwidth}
\caption{\textbf{Comparison of different negative sampling strategies for MIPO.} We consider three different ways of data augmentation: \textbf{(1) reshuffling}, \textbf{(2) random contexts or prompts}, and \textbf{(3) missing contexts}. For personalization, approach (2) pairs rejected $y_r \sim p(y|x, c')$ with chosen $y_c \sim p(y|x,c)$ for the query-context pair $(x, c)$; and approach (3) pairs rejected $y_r \sim p(y|x)$ with chosen $y_c \sim p(y|x, c)$.} 
\label{appendix:rejection_sampling}
\centering
\adjustbox{width=0.9\columnwidth}{
\begin{tabular}{lccccccc}
\toprule
Model & CA & PRISM & MB & GSM & SVAMP & ARC Easy & ARC Challenge  \\
\midrule
\textbf{Llama-3.2-1B-Instruct} & 78 & 72.17 & 79.76 & 22 & 51.67 & 44 & 33.2 \\ 
\quad + Reshuffling & 88.33 (2.02) & 77.84 (3.09) & 84.52  (2.38) & \textbf{29.50 (2.29)} & \textbf{60.11 (0.19)} & 51.87 (2.01) & \textbf{39.53 (3.11)}\\
\quad + Random & 87 (5.57) & 77.32 (1.36) & 89.68 (1.92) & 27.33 (1.26) & 52.22 (1.02) & \textbf{53.47 (2.80)} & 27.33 (1.26)\\
\quad + Missing & \textbf{93.67 (1.26)} & \textbf{80.93 (3.61)} & \textbf{93.26 (0.69)} & - & - & - & - \\ 
\midrule

\textbf{Llama-3.2-3B-Instruct} & 78 & 76.80 & 83.33 & 71 & 78.67 & 80.4 & 68.6 \\ 
\quad + Reshuffling & 86.63 (1.89) & 80.76 (2.84) & 91.27 (1.82) & 70.17 (2.02) & \textbf{78.22 (1.57)} & 85.26 (0.58) & 70.93 (0.70)  \\
\quad + Random & 88 (2.29) & \textbf{83.41 (0.52)} & \textbf{94.84 (1.82)} & \textbf{72.17 (0.29)} & 76.89 (0.77) & \textbf{85.40 (0.69)} & \textbf{72.67 (1.81)} \\
\quad + Missing & \textbf{90.33 (2.08)} & 81.62 (2.08) & 94.45 (2.48) & - & - & - \\
\midrule 

\textbf{Qwen2.5-1.5B-Instruct} & 63.5 & 39.18 & 39.29 & 65.5 & 82.33 & 79 & 63.6 \\
\quad + Reshuffling & 71.83 (1.53) & 50.86 (0.78) & 40.87 (3.44) & \textbf{71 (1.80)} & \textbf{81.67 (2.03)} & \textbf{82.27 (0.81)} & 65.93 (1.70) \\ 
\quad + Random & 68 (1.00) & \textbf{60.31 (1.36)}  & 55.95 (3.57) & 70.33 (1.15) & 81.33 (1.20) & 81.87 (0.31) & \textbf{66.53 (1.10)} \\
\quad + Missing & \textbf{78.83 (2.84)} & 59.79 (1.36) & \textbf{74.6 (3.0)} & - & - & - & - \\ 
\midrule

\textbf{Qwen2.5-3B-Instruct} & 76 & 49.49 & 63.10 & 84.5 & 90.67  & 92 & 79.4  \\
\quad + Reshuffling & 76.17 (1.76) & 56.19 (0.52) & 65.87 (4.51) & \textbf{89.17 (3.75)} & \textbf{91.33 (1.73)} & 90.80 (0.35) & 80.13 (2.01) \\
\quad + Random & \textbf{80.33 (1.76)} & \textbf{60.31 (1.36)} & 70.63 (2.48) & 86.67 (2.25) & \textbf{91.33 (0.33)} & \textbf{91.13 (0.42)} & \textbf{80.80 (0.80)} \\
\quad + Missing & 79.83 (0.29) & 59.79 (1.36) & \textbf{74.60 (0.69)} & - & - &- &- \\ 
\bottomrule
\end{tabular}}
\vspace{-0.2cm}
\end{table}

\begin{table}[H]
% \vspace{-0.5cm}
% \centering
% \begin{minipage}{0.5\textwidth}
\caption{\textbf{Increasing the number of negatives improves MIPO performance.} We trained MIPO with $N = 1, 3, 5, 10$ negatives, where rejected samples are generated using random contexts on Multifaceted Bench. We find that performance generally improves with larger $N$, as the MI lower bound tightens with increasing $N$. We also added random negatives which are random alphanumeric strings of the same lengths as the chosen responses. While the other negatives are drawn from the reference policy's distribution, these random negatives lie outside the policy's support and thus break the MIPO's theoretical assumption, despite providing clear contrastive signals with the chosen responses. Therefore, as expected, MIPO with random negatives performs worse than MIPO with theoretically aligned negatives. Best performance from each row is boldfaced.} 
\label{appendix:n_sampling}
\centering
\adjustbox{width=0.5\columnwidth}{
\begin{tabular}{lccccc}
\toprule
$N$ & Random &  1 & 3  & 5 & 10  \\
\midrule
\textbf{Llama-3.2-1B-Instruct} &  79.8 & 89.7 &  89.3 & 91.7 & \textbf{94.1} \\ 

\textbf{Llama-3.2-3B-Instruct} &  90.5 & 94.8 & \textbf{95.2} & \textbf{95.2} & 92.2 \\ 

\textbf{Qwen2.5-1.5B-Instruct} & 34.5 & 56.0 & 54.8 & 57.1 & \textbf{58.3} \\ 

\textbf{Qwen2.5-3B-Instruct} &  64.3 & 70.6 & 75.0 & \textbf{81.0} & 78.6 \\

\bottomrule
\end{tabular}}
\vspace{-0.2cm}
\end{table}

\begin{table}[H]
\caption{\textbf{Critic performance evaluated using ground-truth human preference labels. We compute the strict and tie-tolerant accuracy of predicting a higher score for the chosen response over the rejected response.}} 
\label{appendix:rm_accuracy}
\centering
\adjustbox{width=0.7\columnwidth}{
\begin{tabular}{lcccccc}  % l = Model, c = 4 data columns, adjust if needed
\toprule
 & \multicolumn{2}{c}{Community Alignment (CA)} & \multicolumn{2}{c}{PRISM} \\
\cmidrule(lr){2-3} \cmidrule(lr){4-5}  % horizontal lines under group headers
Critic model & Strict acc. & Tie-tolerant acc. & Strict acc. & Tie-tolerant acc. \\  % optional sub-headers for each column
\midrule
Llama-3.2-1B-Instruct & \textcolor{red}{\textbf{6}} & 98.5 & \textcolor{red}{\textbf{36.5}} & 76.5 \\
Llama-3.2-3B-Instruct & 54 & 89.5 & 72.5 & 87.5 \\
Llama-3-8B-Instruct & 76 & 92 & 68.5 & 92 \\
Qwen2.5-1.5B-Instruct  & \textbf{\textcolor{red}{27.5}} & 87 & \textcolor{red}{\textbf{42.5}} & 78 \\
Qwen2.5-3B-Instruct  & \textcolor{red}{\textbf{6}} & 100 & 74.5 & 91.5 \\
Qwen2.5-7B-Instruct  & 63.5 & 98.5 & 72.46 & 92.6 \\
Qwen2.5-14B-Instruct  & 63 & 93.5 & 78 & 91.5 \\
\bottomrule
\end{tabular}}
\end{table}

While prior work on RLHF has also observed similarly moderate reward model accuracy of 69.6\% and 72.4\%, the learned policy still achieved substantial performance gains (see Appendix E.2~\citep{ouyang2022traininglanguagemodelsfollow}). However, we suspect that an accuracy below 50\% or unreliable rewards (i.e., the critic always gives 1 or 5 indiscriminately regardless of the response quality) is detrimental to learning and can lead to performance degradation after RLAIF. 

In particular, we observe that Llama-1B on Community Alignment mostly assigns a score of 1 to both chosen and rejected responses; while Qwen-3B mostly assigns a score of 5. In both cases, the critic fails to discriminate between the chosen and rejected responses. 

% \begin{table}[H]
% \caption{\textcolor{blue}{\textbf{Evaluation on ARC-Challenge and Easy datasets using iid rejected samples from $y_r \sim \pi_\text{ref}(y|x'), x' \sim \mathcal X$}. TODO! with hyperparam tuning to check performance?}} 
% \label{appendix:iid_sampling}
% \centering
% \adjustbox{width=0.5\columnwidth}{
% \begin{tabular}{lccc}
% \toprule
% Model & ARC (Easy) & ARC (Challenge) & Avg. \\
% \midrule
% \textbf{Llama-3.2-1B-Instruct}  \\
% \quad +  \\
% \midrule
% \textbf{Llama-3.2-3B-Instruct}  \\
% \quad + \\
% \midrule 
% \textbf{Qwen2.5-1.5B-Instruct}  \\
% \quad + \\
% \midrule
% \textbf{Qwen2.5-3B-Instruct}  \\
% \quad +  \\
% \bottomrule
% \end{tabular}}
% \end{table}

\begin{figure}[h]
% \vspace{-0.2cm}
\begin{center}
\centerline{\includegraphics[width=0.65\textwidth]{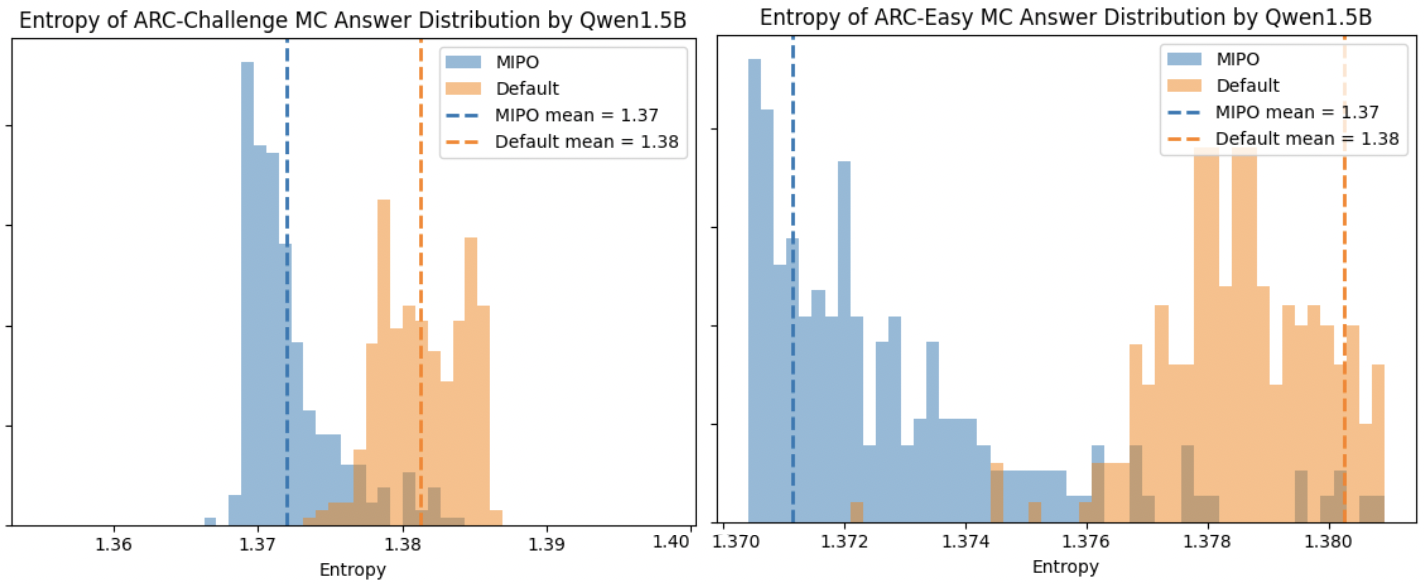}}
\caption{\textbf{Entropy over the MCQ answer choices conditioned on correct model predictions.} In addition to improving accuracy, MIPO (blue) also makes models become more confident about correct predictions compared to the base model (orange) as indicated by the mean and overall distribution shift. The x-axis ranges from mean $\pm$ 1 std.}
\label{fig:mcq_entropy}
\end{center}
% \vspace{-1.5cm}
\end{figure}

\begin{figure}[h]
% \vspace{-0.2cm}
\begin{center}
\centerline{\includegraphics[width=0.7\textwidth]{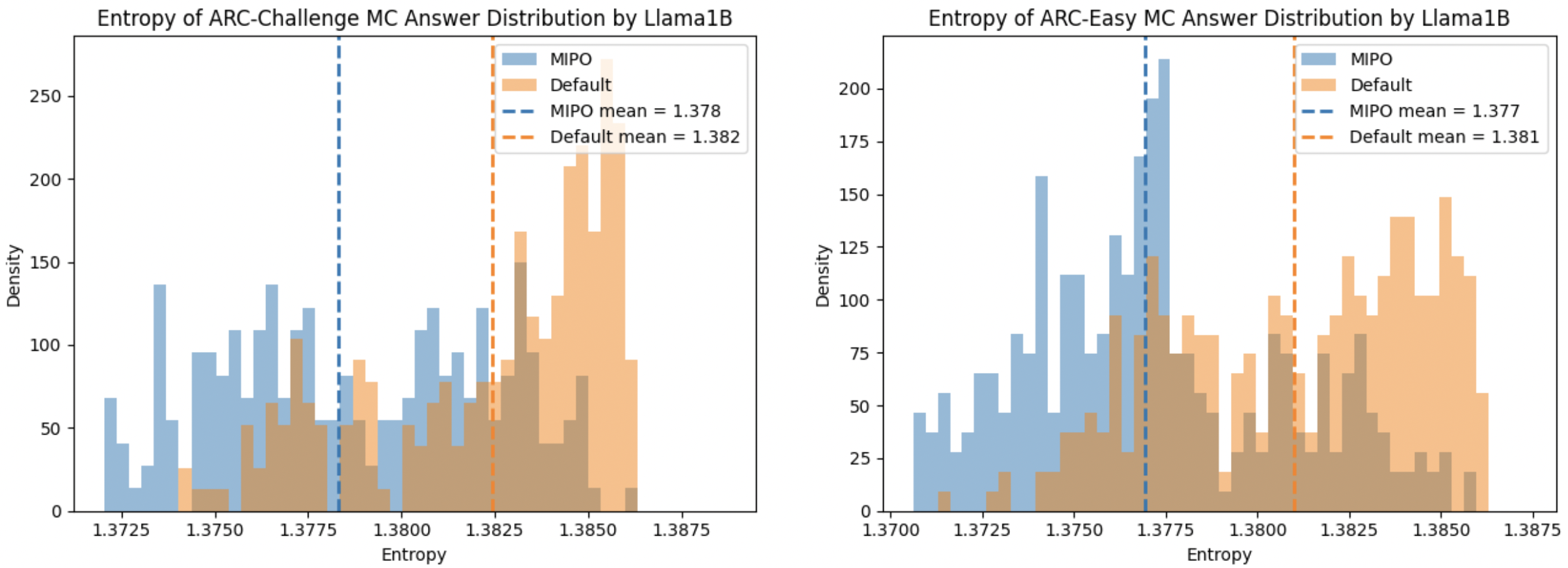}}
\caption{\textbf{Entropy over the 500 MCQ answer choices in ARC conditioned on output correctness.} MIPO-trained models (blue) become more confident about correct answers compared to the base models (orange) as indicated by entropy reduction. The range is truncated to show the mean $\pm$ 2 standard deviation.}
\label{fig:mcq_entropy_llama}
\end{center}
\vspace{-0.2cm}
\end{figure}
\clearpage 

\section{Appendix: MIPO-Personalized Responses}
\label{appendix:personalization_examples}
% \centering
\begin{tcolorbox}[before skip=6pt,
  after skip=6pt,
  colback=white,
  colframe=blue!40,
  coltitle=black,
  colbacktitle=blue!15,
  title= Examples from the Multi-Bench personalization benchmark,
  boxrule=0.5pt,
  arc=2mm
]
\small \textbf{Prompt:} I like answers that show sensitivity and consideration for different people's needs or preferences, especially when discussing things like food. That means being \textbf{\emph{thoughtful about dietary restrictions, cultural tastes, or health-related matters.}} Given the following dialogue, answer to the question. [\textit{Dialogue about people deciding on lunch.}], ``Target": ``No, I think I'll come with you. I'm longing for a nice salad.", ``Question": ``What is or could be the cause of target?" \\ \\
\textbf{Default \#1 (Llama3.2-3B-Instruct)}: In this dialogue, \textcolor{red}{the cause of "target" is likely the coffee shop where they plan to go for lunch.} The speaker, Louise, suggests going to the coffee shop, and the other person agrees to join her, indicating that the coffee shop is the target of their lunch plans. \\ \\
\textbf{MIPO-Personalized \#1 (Llama3.2-3B-Instruct)}: The cause of B's statement could be a few factors: \\ \\ 
\textcolor{blue}{\textbf{1. **Dietary Preferences**:}} B might prefer salads over other types of food, especially if they are health-conscious or follow a specific diet. \\ \\ 
\textcolor{blue}{\textbf{2. **Cultural Tastes**:}} B might have a cultural preference for salads, which aligns with their desire to eat one. \\ \\ 
\textcolor{blue}{\textbf{3. **Health-Related Matters**:}} B might be trying to maintain a healthy lifestyle and see the salad as a better option than a cheeseburger and fries. \\ \\ 
\textcolor{blue}{\textbf{4. **Social Interaction**: B might want to spend more time with A and enjoy a meal together, and a salad might be a suitable choice that both can agree on.}} \\ \\ 
These factors collectively contribute to B's decision to join A for lunch and choose a salad.
\\ \\
\small \textbf{Prompt:} I like it when stories focus on the personalities and emotions of the characters. It helps me connect with them and understand their decisions better. Given the following dialogue, answer to the question. [\textit{Dialogue about people deciding on lunch.}], ``Target": ``No, I think I'll come with you. I'm longing for a nice salad.", ``Question": ``What is or could be the cause of target?" \\ \\
\small \textbf{Default \#2 (Llama3.2-3B-Instruct):} Based on the dialogue, the cause of "I'm longing for a nice salad" could be that \textcolor{red}{the person (B) is not getting enough variety in their diet or is craving something healthy, or possibly they are trying to maintain a healthy lifestyle.}
\\ \\ 
\small \textbf{MIPO-Personalized \#2 (Llama3.2-3B-Instruct):} Based on the dialogue, it seems that the cause of B's decision to accompany A to the coffee shop and order a salad could be a combination of several factors. Here are a few possibilities:\\ \\ 
\textcolor{blue}{\textbf{1. **B's desire for a break from their usual routine**:}} By saying "I usually bring my lunch from home and eat at my desk," B implies that they often spend most of their day working alone, which might be isolating. The invitation to go out and get lunch together could be a way for B to break up the monotony and socialize with A. \\ \\ 
\textcolor{blue}{\textbf{2. **B's emotional state**:}} The phrase "I'm longing for a nice salad" suggests that B might be feeling a bit down or stressed, and the thought of eating a salad at a coffee shop is a way for them to treat themselves and lift their mood. \\ \\ 
\textcolor{blue}{\textbf{3. **B's desire to connect with A**:}} By agreeing to go out with A, B might be seeking a chance to bond with their friend and have some quality time together. This could be a way for B to show that they value their relationship with A and want to spend time together. \\ \\ 
\textcolor{blue}{\textbf{4. **B's curiosity about A's preferences**:}} B's question about where A would like to eat and their subsequent decision to join A might also be driven by a desire to understand A's tastes and preferences. By trying the coffee shop with A, B can get a sense of what A enjoys and might even discover new things about their friend. \\ \\ 
Overall, B's decision to accompany A to the coffee shop seems to be motivated by a mix of practicality, emotional needs, and a desire to connect with their friend.
% Answers to the first prompt
% \textbf{Default \# 2 (Qwen2.5-Qwen3B-Instruct):} The target statement "Do you often have lunch at this coffee shop? I\'m longing for a nice salad." suggests that the person (B) has a preference for salads and is not used to eating at the coffee shop. This implies that B might have dietary restrictions, such as being vegetarian, vegan, or having a preference for low-calorie options, which might not be available in the coffee shop. Alternatively, B might simply prefer salads due to personal taste or health reasons. The context also indicates that B usually brings their own lunch from home, further supporting the idea that they might have specific dietary needs or preferences that differ from what the coffee shop offers.
% \\ \\ 
% \textbf{MIPO-Personalized \# 2 (Qwen2.5-Qwen3B-Instruct):} In this dialogue, the target statement "I'm longing for a nice salad" shows sensitivity and consideration for Louise's preferences and needs. By mentioning her desire for a salad, Louise is taking into account potential dietary restrictions or health considerations that might not align with a cheeseburger and fries. This response also respects cultural tastes and personal choices, indicating that she values a balanced meal that includes a salad, which is a common preference for those mindful of their health or looking for a lighter option. \textcolor{blue}{\textbf{This approach demonstrates thoughtful consideration for both her own needs and potentially your need to accommodate any dietary restrictions or preferences you might have.}}
\end{tcolorbox}
\clearpage 

\begin{tcolorbox}[before skip=6pt,
  after skip=6pt,
  colback=white,
  colframe=blue!40,
  coltitle=black,
  colbacktitle=blue!15,
  title= Examples from the Multi-Bench personalization benchmark,
  boxrule=0.5pt,
  arc=2mm
]
\small \textbf{Prompt \#3:} I’m pretty familiar with casual dining and how people tend to interact in social eating situations, so you don’t have to explain the basics to me. Instead, include things that are more specific or relatable to the vibe of going out with friends or coworkers. Given the following dialogue, answer to the question. [\textit{Dialogue about people deciding on lunch.}], ``Target": ``No, I think I'll come with you. I'm longing for a nice salad.", ``Question": ``What is or could be the cause of target?" \\ \\
\small \textbf{Default \#3 (Llama3.2-3B-Instruct):} The cause of "Target" in this dialogue is that Louise is suggesting that B should join her at the coffee shop, implying that the coffee shop is a suitable place for a casual lunch outing.
\\ \\ 
\small \textbf{MIPO-Personalized \#3 (Llama3.2-3B-Instruct):} In this dialogue, the phrase "I'm longing for a nice salad" is a idiomatic expression that suggests B is craving a salad, but it's not just a simple desire for a healthy meal. The word "longing" implies a stronger emotional attachment or nostalgia for a salad, possibly because it's a comfort food or a favorite of theirs. It could also be that \textcolor{blue}{they're looking for a break from their usual routine of bringing lunch from home and eating at their desk, and the idea of going out to a coffee shop and trying a new food option is appealing to them. The phrase is more about the emotional connection they have with the idea of eating out and trying something different, rather than just a physical need for a salad.} \\ \\
\textbf{Prompt \#4:} I like when you go deeper into the meaning behind what people say. Social cues and hidden intentions make conversations more interesting, so I’d prefer if you unpack those layers instead of just sticking to surface-level meanings. Given the following dialogue, answer to the question.  \\ \\ 
\textbf{Default \#4 (Llama3.2-3B-Instruct):} Let's unpack the layers of this dialogue to understand the underlying intentions and emotions behind B's response.

At first glance, B's statement "No, I think I'll come with you. I'm longing for a nice salad" seems straightforward, but upon closer analysis, it reveals a more complex emotional landscape.

The phrase "I'm longing for a nice salad" is a polite way of saying that B is not actually hungry for a salad, but rather, they're seeking a distraction or a break from their current situation. The word "longing" implies a sense of yearning or craving, which is often associated with emotional or psychological needs rather than physical hunger.

In this context, B's statement can be seen as a way of avoiding the possibility of eating at the coffee shop, which might be a more casual or social setting that B is not comfortable with. By saying they're "longing for a salad," B is subtly hinting that they're not really interested in eating at the coffee shop, but rather, they're seeking a more solitary or low-key experience.

Furthermore, B's decision to "come with" A, despite not being hungry, suggests that they're seeking companionship or a sense of connection. This could be a way of avoiding feelings of loneliness or isolation, or perhaps they're seeking to create a sense of obligation or responsibility towards A.

Overall, B's response is a masterful example of social politeness, where they're using a seemingly innocuous statement to convey a more complex emotional landscape. By unpacking the layers of this dialogue, we can see that B's true intentions are not just about eating a salad, but about seeking connection, avoiding social situations, and managing their emotions. \\ \\ 
\textbf{MIPO-Personalized \#4 (Llama3.2-3B-Instruct):} Let's dive deeper into the conversation to uncover the underlying intentions and emotions behind B's response.

When B says, "No, I think I'll come with you. I'm longing for a nice salad," there are several layers to this statement. On the surface, it seems like B is simply agreeing to join A for lunch and expressing a desire for a salad. However, the word "longing" suggests a deeper emotional connection.

The word "longing" typically implies a strong desire or yearning for something, often accompanied by a sense of nostalgia or wistfulness. In this context, it's possible that B is not just craving a salad, but also seeking a sense of comfort, relaxation, or escape from their current situation.

By saying they usually bring their lunch from home and eat at their desk, B may be hinting that their work environment is not conducive to taking a break or enjoying a leisurely lunch. This could be due to a stressful job, a busy schedule, or a sense of obligation. By suggesting they'll join A for lunch, B might be seeking a brief respite from their work-related pressures and a chance to recharge.

Furthermore, B's decision to join A for lunch could also be motivated by a desire for social interaction or human connection. In today's work culture, it's common for people to spend most of their day staring at screens and interacting with colleagues only through digital means. By joining A for lunch, B may be seeking a sense of community and social validation.

\textcolor{blue}{Additionally, B's response could also be influenced by a sense of obligation or politeness. They might feel obligated to spend time with A, or they might be trying to be friendly and courteous. This could be a way of maintaining a positive relationship or avoiding conflict. \emph{\# This possibility does not appear in the default response.}}

\textcolor{blue}{Overall, B's statement "No, I think I'll come with you. I'm longing for a nice salad" suggests that they're seeking a combination of comfort, social connection, and a break from their work-related stress. By joining A for lunch, B is not just getting a salad, but also a chance to momentarily escape their daily routine and connect with someone on a more personal level.}
\end{tcolorbox}

\section{Appendix: Datasets}\label{appendix:datasets}
\subsection{Standard LLM Benchmarks on Math and MCQ}\begin{itemize}
\item \textbf{{\includegraphics[height=1em]{math_clipart.png}} GSM8k~\citep{cobbe2021trainingverifierssolvemath}} is a standard math benchmark for grade-school-level problems. We sampled 5000 problems from the training set and 200 problems from the evaluation set, and used the fixed 8-shot from ~\citet{geng2025deltalearninghypothesispreference}, which are included below.  \textbf{SVAMP~\citep{patel2021nlpmodelsreallyable}} is also a math problem benchmark of similar difficulty level as GSM~\citep{huang2025mathperturbbenchmarkingllmsmath}. Since SVAMP is too small to use for both training and evaluation, we sample 700 problems from SVAMP and 700 from GSM8k for training data, and evaluate on the remaining 300 SVAMP problems. Models output a numeric answer after \texttt{\#\#\#\#} for easy parsing.

% \item \textbf{{\includegraphics[height=1em]{alphabet_clipart.png}}  MMLU~\citep{hendrycks2021measuringmassivemultitasklanguage}} is a dataset for knowledge-intensive question answering across 57 domains, including US history, computer science, and law. We sample 5,000 samples from the training and evaluate on 200 samples from the test split. Models output reasoning followed by their selection of the correct answer as \texttt{"The answer is: <X>"}.

\item \textbf{{\includegraphics[height=1em]{alphabet_clipart.png}} AI2 Reasoning Challenge (ARC)~\citep{clark2018thinksolvedquestionanswering}} contains 7,787 MCQs targeted at grade-school level science that are split into Easy and Challenge based on their problem difficulty, and is one of the reasoning benchmarks included in OLMES for standard LLM evaluation~\citep{gu2025olmesstandardlanguagemodel}. We use the train split for each problem subset and evaluate on 500 samples from each test set.
\end{itemize}

\subsection{Personalized Instruction-Following Tasks}
\begin{tcolorbox}[before skip=6pt,
  after skip=6pt,
  colback=white,
  colframe=blue!40,
  coltitle=black,
  colbacktitle=blue!15,
  title=GPT-4o instruction for extracting user contexts from real-user datasets,
  boxrule=0.5pt,
  arc=2mm
]
\textbf{System prompt:} You are a helpful assistant that infers a user's preferences based on their selections from a previous conversation example. Write personalization instructions based on these inferred preferences. The instructions should be general enough to apply to future conversations on different topics and formatted as a JSON object. In JSON, all keys and string values must be enclosed in double quotes ("). For example, \texttt{"key name"}: \texttt{"value"} is valid, but \texttt{key name}: "value" or \texttt{'key name'}: 'value' are not. Each key should use snake case, summarizing the corresponding instruction in one or two words, and the value should be detailed instructions of no more than two sentences. Avoid using any topic specific instructions and focus on the user's general preferences.\\ \\
\textbf{User's prompt:} , "Response options": f"Option A: Option B: ", "User's preferred response" .
\end{tcolorbox}

Then we aggregate all the values as user-contexts to build the context set $\mathcal C$.

\begin{tcolorbox}[before skip=6pt,
  after skip=6pt,
  colback=white,
  colframe=blue!40,
  coltitle=black,
  colbacktitle=blue!15,
  title=GPT-4o instruction for converting system messages into user messages for Multi-Bench,
  boxrule=0.5pt,
  arc=2mm
]

\textbf{System prompt:} You are a helpful and creative assistant tasked with generating different user preference descriptions.
\\ \\ 
example = json.dumps(\{"dimension": "style", "sub-dimension": "conciseness","preference": "straight-to-the-point narratives","explanation": "This preference favors narratives that are concise and avoid complex language. It values brevity and the clarity of simple, intuitive examples.", "user-statement": "I prefer simple, intuitive examples and dislike long, hard-to-understand explanations that take forever to read.\})
\\ \\ 
new-example = json.dumps(\{"dimension": dimension, "sub-dimension": subdimension, "preference": pref, "explanation": description, "user-statement": "fill this with your response"\})
\\ \\ 
\textbf{User message:} Given the following information about the user's preferences, rewrite it as something the user would say to the assistant. Output the response in JSON format. For example: \{example\}.
\\ \\ 
Here's a new example: \{new example\}.
\end{tcolorbox}

\begin{tcolorbox}[before skip=6pt,
  after skip=6pt,
  colback=white,
  colframe=blue!40,
  coltitle=black,
  colbacktitle=blue!15,
  title=GPT-4o instruction for generating reward model scoring rubric,
  boxrule=0.5pt,
  arc=2mm
]
\textbf{System prompt:} You are a helpful and creative assistant tasked with generating different user preference descriptions. \\ \\
example = json.dumps(\{"dimension": "style", "sub-dimension": "conciseness","preference": "straight-to-the-point narratives","explanation": "This preference favors narratives that are concise and avoid complex language. It values brevity and the clarity of simple, intuitive examples.", "user-statement": "I prefer simple, intuitive examples and dislike long, hard-to-understand explanations that take forever to read."\}) \\ \\
new-example = json.dumps(\{"dimension": dimension, "sub-dimension": subdimension, "preference": pref, "explanation": description, "user-statement": "fill this with your response"\}) \\ \\
\textbf{User message:} Given the following information about the user's preferences, rewrite it as something the user would say to the assistant. Output the response in JSON format. For example: \{example\}. Here's a new example: \{new-example\}. \\ \\
\textbf{Rubric example \# 1:} \{'criteria': 'Does the response demonstrate an intermediate understanding of narrative structures and effectively use dialogue and action instead of adjectives and adverbs?', 'score-descriptions': \{'1': 'The response heavily relies on adjectives and adverbs, showing minimal use of dialogue or action, indicating a poor grasp of narrative structures.', '2': 'The response uses some dialogue and action, but still predominantly depends on adjectives and adverbs, reflecting only a basic understanding of narrative structures.', '3': 'The response balances the use of dialogue, action, and descriptive language, suggesting a moderate understanding of narrative structures.', '4': 'The response predominantly uses dialogue and action to convey details, with minimal reliance on adjectives and adverbs, displaying a good understanding of narrative structures.', '5': 'The response skillfully employs dialogue and action to convey details, completely avoiding adjectives and adverbs, demonstrating an excellent mastery of intermediate narrative structures.'\}\}
\\ \\ 
\textbf{Rubric example \# 2:} \{'criteria': 'Does the model efficiently convey a narrative using storytelling techniques that minimize the use of adjectives and adverbs, focusing instead on dialogue, narrative pacing, and character actions?', 'score-descriptions': \{'1': 'The response heavily relies on adjectives and adverbs with minimal or no use of suggested storytelling techniques.', '2': 'The response includes some use of dialogue, pacing, or character actions but still predominantly relies on adjectives and adverbs.', '3': 'The response balances the use of adjectives and adverbs with effective storytelling techniques like dialogue and character actions, though not optimally.', '4': 'The response skillfully uses storytelling techniques such as dialogue, pacing, and character actions with limited reliance on adjectives and adverbs, enhancing narrative efficiency.', '5': 'The response excellently prioritizes narrative efficiency through masterful use of dialogue, narrative pacing, and character actions, virtually eliminating unnecessary adjectival and adverbial descriptions.'\}\}
\\ \\ 
\textbf{Rubric example \# 3:} \{'criteria': 'Does the response exhibit linguistic creativity by effectively communicating without using adjectives and adverbs, while maintaining accuracy and clarity in the message?', 'score-descriptions': \{'1': 'The response lacks creativity and clarity, with incorrect or inappropriate use of language. Adjectives and adverbs are used, detracting from the effectiveness of the communication.', '2': 'The response shows minimal creativity, often relying on adjectives and adverbs. It maintains basic accuracy but fails to enrich the narrative or ensure clarity.', '3': 'The response is somewhat creative, occasionally omitting adjectives and adverbs. While generally accurate, it lacks consistent clarity and richness in narrative.', '4': 'The response is creatively composed with rare use of adjectives and adverbs, maintaining good accuracy and clarity. It could, however, further enrich the narrative.', '5': 'The response exemplifies outstanding linguistic creativity with zero use of adjectives and adverbs, delivering a clear, accurate, and rich narrative.'\}\} \\ \\
\textbf{User message:} Given the following user's preference, generate a JSON scoring rubric with criteria and score-descriptions. Here are some examples:

Example 1: \{json.dumps(rubric-example1)\}

Example 2: \{json.dumps(rubric-example2)\}

Example 3: \{json.dumps(rubric-example3)\}

New preference: \{pref\}.
\end{tcolorbox}

\begin{tcolorbox}[before skip=6pt,
  after skip=6pt,
  colback=white,
  colframe=blue!40,
  coltitle=black,
  colbacktitle=blue!15,
  title=RLAIF scoring instruction,
  boxrule=0.5pt,
  arc=2mm
]

You are a helpful assistant, that evaluates another assistant's response based on how well the response is personalized to this user based on their stated preference. Give a score between 1 and 5, 1 if the response is general, and 5 if the response sufficiently addresses the user's preference. Only respond with a number between 1 and 5 and do not provide an explanation.
\\ \\
User preference: \{context\}
\\ \\
User: \{user's query\} 
\\ \\
Assistant: \{model output\}
\end{tcolorbox}

\begin{tcolorbox}[before skip=6pt,
  after skip=6pt,
  colback=white,
  colframe=blue!40,
  coltitle=black,
  colbacktitle=blue!15,
  title=Reward evaluation instruction,
  boxrule=0.5pt,
  arc=2mm
]\label{appendix:reward_prompt}
\textcolor{blue}{\#\# For each item in the rubric, we run the scorer and aggregate the scores; Qwen models are evaluated by Qwen-2.5-14B-Instruct, Llama models are evaluated by Llama-3-8B-Instruct to avoid any cross-model inconsistencies} \\ \\
You are a helpful assistant, that evaluates another assistant's response based on how well the response is personalized to this user based on their stated preference. Give a score between 1 and 5, 1 if the response is general, and 5 if the response sufficiently addresses the user's preference. Only respond with a number between 1 and 5 and do not provide an explanation.
\\ \\
User: \{user's query\} 
\\ \\
Assistant: \{model output\}
\\ \\ 
User's preference: \{itemized instruction for personalization (e.g., user's preferences and circumstances)\}
\end{tcolorbox}

\begin{tcolorbox}[before skip=6pt,
  after skip=6pt,
  colback=white,
  colframe=blue!40,
  coltitle=black,
  colbacktitle=blue!15,
  title=Example user query with and without in-context user preferences,
  boxrule=0.5pt,
  arc=2mm
]
\textbf{Vanilla user query: } Give me short 2-3 sentence answers to the following prompt: Draft a description of my art studio, highlighting the projects I'm working on, and inviting friends and collectors to visit and give feedback.
 \\ \\ 
\textbf{Query with user-specific preferences/attributes in context: } Give me short 2-3 sentence answers to the following prompt: Draft a description of my art studio, highlighting the projects I'm working on, and inviting friends and collectors to visit and give feedback. Please respect the following requirements made by the user: \emph{Use a warm and engaging tone that conveys enthusiasm and passion. Provide vivid and immersive descriptions to create a rich and evocative experience. Encourage interaction and collaboration by emphasizing the importance of community and shared ideas. Extend open and welcoming invitations while showing genuine appreciation for feedback and participation. Incorporate sensory details like sounds, scents, and visual aesthetics to enhance the experience.}
\end{tcolorbox}

\begin{tcolorbox}[before skip=6pt,
  after skip=6pt,
  colback=white,
  colframe=blue!40,
  coltitle=black,
  colbacktitle=blue!15,
  title=Prompt used for self-improvement with SFT,
  boxrule=0.5pt,
  arc=2mm
]
\label{appendix:sft_revision}
\textbf{Prompt used for self-revision: } You are a helpful assistant. Modify the assistant's message to maek sure the response is personalized to this user based on the information available about them. You will be given an example of a good response and a bad response to the same prompt. 
 \\ \\ 
\texttt{user\_context + prompt} 
\\  \\
Good assistant's response: \texttt{good\_example} \\ \\ Bad assistant's response: \texttt{bad\_example}. \\ \\ Rewrite the good assistant's response to make it better.
\end{tcolorbox}

\begin{tcolorbox}[before skip=6pt,
  after skip=6pt,
  colback=white,
  colframe=blue!40,
  coltitle=black,
  colbacktitle=blue!15,
  title=LLM-judge evaluation prompt for 70B experiments,
  boxrule=0.5pt,
  arc=2mm
]
\label{appendix:binary_evaluation}
\# We repeat the same query twice by switching the place of \texttt{response\_a} and \texttt{response\_b} to avoid the model's position bias. \\ \\ 
\textbf{System message: } You are a fair judge. Use the following rubric to decide whether response\_a or response\_b is better. Respond with response\_a or response\_b. Do not include any additional text. \\ \\ 
\textbf{User message: } \texttt{criteria} \\ \\ Response\_a: \texttt{response\_a} \\ \\ Response\_b: \texttt{response\_b} \\ \\ 
\# Criteria change for different users. For example, $c_1 = $ \textit{``Does the response employ rich, descriptive language to vividly depict the scene or subject, enhancing storytelling and reader engagement? Use the following scoring rubric: \{"1": "The response is bland and lacks descriptive elements, failing to paint a vivid picture or evoke imagery.", "2": "The response includes some descriptive language but fails to consistently evoke vivid imagery or fully engage the reader.", "3": "The response uses descriptive language fairly well, creating a somewhat vivid picture, though it could be more engaging or detailed.", "4": "The response effectively uses rich, descriptive language, vividly depicting the scene or subject with minor lapses in detail or engagement.", "5": "The response excellently employs rich, descriptive language, vividly and compellingly drawing the reader into the scene or subject with great detail and emotional impact."\}"}, and $c_2 = $ \textit{``Does the model demonstrate an expert-level understanding of forensic techniques as per the user\'s preference? Use the following scoring rubric: \{"1": "The model\'s response shows no understanding of forensic techniques and fails to align with expert-level insights.", "2": "The model\'s response demonstrates a basic understanding, but lacks depth and largely misses the expert-level insights expected.", "3": "The model\'s response reflects a moderate understanding of forensic techniques, but the expert-level depth and detail are inconsistent.", "4": "The model\'s response is mostly aligned with expert-level understanding, containing well-integrated forensic insights with minor inaccuracies.", "5": "The model\'s response perfectly aligns with an expert-level understanding, showcasing deep insights and comprehensive knowledge of forensic techniques."\}"}
\end{tcolorbox}

\subsection{Verifiable Domains}
\begin{tcolorbox}[before skip=6pt,
  after skip=6pt,
  colback=white,
  colframe=blue!40,
  coltitle=black,
  colbacktitle=blue!15,
  title=8-shot examples used in GSM8k and SVAMP,
  boxrule=0.5pt,
  arc=2mm
]
\textcolor{blue}{Few-shot examples are from \texttt{FEWSHOT-SOURCES["STD:GSM8k"]} of \href{https://github.com/allenai/olmes/blob/main/oe_eval/tasks/fewshot_sources.py}{the OLMES repository}.} \\ \\
FEWSHOT SOURCES = [ \\
    \{
        "question": "There are 15 trees in the grove. Grove workers will plant trees in the grove today. After they are done, there will be 21 trees. How many trees did the grove workers plant today?",
        "answer": "There are 15 trees originally. Then there were 21 trees after some more were planted. So there must have been 21 - 15 = 6. So the answer is 6.",
        "short answer": "6",
    \}, \\ \\ 
    \{
        "question": "If there are 3 cars in the parking lot and 2 more cars arrive, how many cars are in the parking lot?",
        "answer": "There are originally 3 cars. 2 more cars arrive. 3 + 2 = 5. So the answer is 5.",
        "short answer": "5",
    \}, \\ \\ 
    \{
        "question": "Leah had 32 chocolates and her sister had 42. If they ate 35, how many pieces do they have left in total?",
        "answer": "Originally, Leah had 32 chocolates. Her sister had 42. So in total they had 32 + 42 = 74. After eating 35, they had 74 - 35 = 39. So the answer is 39.",
        "short answer": "39",
    \}, \\ \\
    \{
        "question": "Jason had 20 lollipops. He gave Denny some lollipops. Now Jason has 12 lollipops. How many lollipops did Jason give to Denny?",
        "answer": "Jason started with 20 lollipops. Then he had 12 after giving some to Denny. So he gave Denny 20 - 12 = 8. So the answer is 8.",
        "short answer": "8",
    \}, \\ \\ 
    \{
        "question": "Shawn has five toys. For Christmas, he got two toys each from his mom and dad. How many toys does he have now?",
        "answer": "Shawn started with 5 toys. If he got 2 toys each from his mom and dad, then that is 4 more toys. 5 + 4 = 9. So the answer is 9.",
        "short answer": "9",
    \}, \\ \\ 
    \{
        "question": "There were nine computers in the server room. Five more computers were installed each day, from monday to thursday. How many computers are now in the server room?",
        "answer": "There were originally 9 computers. For each of 4 days, 5 more computers were added. So 5 * 4 = 20 computers were added. 9 + 20 is 29. So the answer is 29.",
        "short answer": "29",
    \}, \\ \\ 
    \{
        "question": "Michael had 58 golf balls. On tuesday, he lost 23 golf balls. On wednesday, he lost 2 more. How many golf balls did he have at the end of wednesday?",
        "answer": "Michael started with 58 golf balls. After losing 23 on tuesday, he had 58 - 23 = 35. After losing 2 more, he had 35 - 2 = 33 golf balls. So the answer is 33.",
        "short answer": "33",
    \}, \\ \\ 
    \{
        "question": "Olivia has \$23. She bought five bagels for \$3 each. How much money does she have left?",
        "answer": "Olivia had 23 dollars. 5 bagels for 3 dollars each will be 5 x 3 = 15 dollars. So she has 23 - 15 dollars left. 23 - 15 is 8. So the answer is 8.",
        "short answer": "8",
    \}, \\ \\ 
] \\ \\ 
examples =  [f"Example {i+1}: " + f"Question: {x['question']} Answer: {x['answer']} \#\#\#\# {x['short answer']}" for i, x in enumerate(FEWSHOT SOURCES)] \\ \\ 
\textbf{User message:} You are a helpful math assistant. Solve the following problem step-by-step and give the final answer after \#\#\#\#. + examples
\end{tcolorbox}
\clearpage
\section{Appendix: Training Details}\label{appendix:training}

- Our code is available at \url{https://github.com/nam630/mutual_information_preference_optimization}. We used two H200 GPUs for most model training, except for PPO on 7B models, which is run on four H200 GPUs. 

- All results, except for Qwen-7B models, are based on three random seeds and are reported with the mean and standard deviation.

- DPO is trained using the DPO implementation of OpenRLHF~\citep{hu2024openrlhfeasytousescalablehighperformance} with the following hyperparameters (we additionally selected the best learning rate from sweeping: \{1e-6, 5e-7, 1e-7\}): 

\begin{description}[leftmargin=3cm]
  \item[\texttt{train\_batch\_size}] 4
  \item[\texttt{micro\_train\_batch\_size}] 1
  \item[\texttt{bf16}] enabled
  \item[\texttt{learning\_rate}] 1e-7
  \item[\texttt{beta}] 0.1
\end{description}

- All models are trained with 1 epoch, except on Multi-Bench, we observe that some models achieve low training accuracy and training for 2--3 epochs on the same data helps improve performance.

- SFT is trained using the SFT implementation of OpenRLHF~\citep{hu2024openrlhfeasytousescalablehighperformance} with the following hyperparameters (similarly as above, we selected the best learning rate from sweeping: \{1e-6, 5e-7, 1e-7\}):
\begin{description}[leftmargin=3cm]
  \item[\texttt{train\_batch\_size}] 4
  \item[\texttt{micro\_train\_batch\_size}] 1
  \item[\texttt{bf16}] enabled
  \item[\texttt{learning\_rate}] 1e-7
\end{description}

- RLAIF and RLVR are trained using the PPO implementation of OpenRLHF~\citep{hu2024openrlhfeasytousescalablehighperformance} with the following hyperparameters: 
\begin{description}[leftmargin=2cm]
  \item[\texttt{micro\_train\_batch\_size}] 1
  \item[\texttt{train\_batch\_size}] 4
  \item[\texttt{micro\_rollout\_batch\_size}] 2
  \item[\texttt{rollout\_batch\_size}] 8
  % \item[\texttt{max\_epochs}] 1
  % \item[\texttt{prompt\_max\_len}] 15360
  % \item[\texttt{generate\_max\_len}] 256
  % \item[\texttt{zero\_stage}] 2
  \item[\texttt{bf16}] enabled
  \item[\texttt{actor\_learning\_rate}] 5e-7
  \item[\texttt{critic\_learning\_rate}] 9e-6
  \item[\texttt{init\_kl\_coef}] 0.001
  % \item[\texttt{apply\_chat\_template}] enabled
  \item[\texttt{normalize\_reward}]
  % \item[\texttt{flash\_attn}] enabled
\end{description}

\clearpage 

\clearpage

\end{document}